\documentclass[10pt,compsoc]{IEEEtran}
\usepackage{babel}
\usepackage{lipsum}
\usepackage{multirow}
\usepackage{booktabs}
\usepackage{makecell}
\usepackage{tabularx}

\usepackage{amsmath}
\usepackage{amssymb}

\usepackage{enumitem}

\usepackage{graphicx}
\usepackage{subfigure}
\usepackage{epstopdf}
\usepackage{epsfig}
\usepackage{xcolor}
\newcommand{\revised}[1]{\textcolor{black}{{#1}}} % Use this definition to see the revise in blue

\begin{document}
\title{Sentence Bag Graph Formulation for Biomedical Distant Supervision Relation Extraction}

\author{Hao~Zhang,
        Yang~Liu,
        Xiaoyan~Liu,
        Tianming~Liang,
        Gaurav~Sharma,~\IEEEmembership{Fellow, IEEE},
        Liang~Xue,
        and~Maozu~Guo% <-this % stops a space
        
\IEEEcompsocitemizethanks
{\IEEEcompsocthanksitem H. Zhang, Y. Liu, X. Liu, T. Liang are with the School of Computer Science and Technology, Harbin Institute of Technology, Harbin 150001, China.
Email: zhanghao2020@stu.hit.edu.cn, and liuyang, liuxiaoyan, liangtianming@hit.edu.cn.

\IEEEcompsocthanksitem G. Sharma is with the Department of Electrical and Computer Engineering, University of Rochester, Rochester, NY 14627, USA.
Email: gaurav.sharma@rochester.edu.

\IEEEcompsocthanksitem L. Xue is with the BYERING.com, HangZhou 310000, China. 
Email: xueliang.xl@byering.com.

\IEEEcompsocthanksitem M. Guo is with the School of Electrical and Information Engineering, Beijing University of Civil Engineering and Architecture, Beijing 100044, China, and also with the Beijing Key Laboratory of Intelligent Processing for Building Big Data, Beijing University of Civil Engineering and Architecture, Beijing 100044, China.
Email: guomaozu@bucea.edu.cn.

\IEEEcompsocthanksitem Corresponding author: Yang Liu and Maozu Guo.}% <-this % stops an unwanted space
}
% \thanks{Manuscript received April 19, 2005; revised August 26, 2015.}}


\markboth{}%
{Shell \MakeLowercase{\textit{et al.}}: Sentence Bag Graph Formulation for Biomedical Distant Supervision Relation Extraction}



\IEEEtitleabstractindextext{%

 \begin{abstract}
   We introduce a novel graph-based framework for alleviating key challenges in distantly-supervised relation extraction and demonstrate its effectiveness in the challenging and important domain of biomedical data. 
   Specifically, we propose a graph view of sentence bags referring to an entity pair, which enables message-passing based aggregation of information related to the entity pair over the sentence bag. 
   The proposed framework alleviates the common problem of noisy labeling in distantly supervised relation \revised{extraction and also} effectively incorporates inter-dependencies between sentences within a bag. 
   Extensive experiments on two large-scale biomedical relation datasets \revised{and the widely utilized NYT dataset} demonstrate that our proposed framework significantly outperforms the state-of-the-art methods for biomedical distant supervision relation extraction \revised{while also providing excellent performance for relation extraction in the general text mining domain}.
\end{abstract}

}

\maketitle

\IEEEdisplaynontitleabstractindextext

% \fi
\IEEEpeerreviewmaketitle

\IEEEraisesectionheading{\section{Introduction}\label{sec:introduction}}

\IEEEPARstart{W}{ith} the continuous development of biomedical research in recent years, a vast amount of biomedical literature is available online, containing valuable healthcare and biomedical data. 
Biomedical relation extraction seeks to automatically extract relations between pairs of biomedical entities mentioned in the literature text through advanced natural language processing techniques, and is beneficial for downstream knowledge-driven biomedical research.
Over the past years, although great efforts have been made in biomedical relation extraction, their impact has been limited by the availability of human-annotated data, whose scale is constrained by the effort required from skilled biomedical scientists and linguistic experts for annotation.\par

Distantly supervised relation extraction (DSRE)~\cite{mintz2009distant} has been proposed as a way to alleviate the problem of inadequate labeled data. DSRE can automatically generate large scale labeled training dataset by aligning entities in texts with the entities in knowledge bases (KBs). 
The distant supervision (DS) assumption is that if a pair of entities has a relation in the KBs, then all sentences that mention the pair of entities will express this relation. 
Obviously, the DSRE assumption is too strong and, therefore, suffers from the problem of noisy labeling. 
For instance, owing to the biomedical relation fact “Abscess (disease condition) - May be treated by – Metronidazole (drug)” in KBs, the sentence “This unique case serves to document that new abscess may develop in the course of metronidazole therapy and illustrates the value of serial hepatoscanning in such patients” will be regarded as an active instance of the relation, although this sentence does not express the relation “may be treated by”.\par

%\IEEEpubidadjcol

Riedel et al.~\cite{2010Modeling} proposed the use of multi-instance learning (MIL) to address the noisy data problem in DSRE and relaxed the DS assumption to the expressed-at-least-once assumption. 
Under the relaxed assumption, if a entity pair participates in a relation, at least one sentence \revised{in the sentence bag} which mentions this entity pair expresses the relation.
In recent years, some work adopted MIL to reduce the influence of label noise via attention mechanisms~\cite{2015Distant,2016Neural,2017Distant,2018Improving,2019batt} and by integrating external knowledge graph information~\cite{2018Label,2022liang}. 
Li et al.~\cite{2019Entity} transform the entity-relation extraction task to a multi-turn question-answering task, and show that the methods of question-answering~\cite{2021robust} and related attention mechanisms~\cite{2017bidirectional,2017coattention} demonstrate great potential for the relation extraction task. 
Compared with the methods using external knowledge, coupling the information of sentence and its corresponding query can explicitly detect the key words in the sentence that are more relevant to the relation between the pair of entities without complex processing of external knowledge. 
We attempt to explore an attention mechanism combined with query to \revised{more effectively} alleviate the noisy labeling problem in DSRE.
Moreover, the attention mechanism will not just filter out a fixed number of important words or discard noise words, because the noisy part of biomedical sentences may also express objective facts between \revised{an} entity pair, which can be viewed as valuable biomedical background knowledge and utilized for the subsequent computation of relevance between sentences.
We obtain a new representation of the complete sentence with less noise and higher quality via the attention.
\par

In the field of biomedical relation extraction, some works are devoted to distant supervision research~\cite{2017Extracting, 2021Distantly+RL}. 
However, the number of relations and instances is limited in the datasets adopted in these works, and do not reflect the large number of relations mentioned in the biomedical literature~\cite{2008textmining}. 
Xing et al.~\cite{2019BioRel} proposed BioRel, the largest domain-general biomedical dataset for DSRE so far, based on Unified Medical Language System (UMLS)~\cite{2004umls} and MEDLINE. 
Due to the rapid growth of biomedical corpora and literature, there will be more sentences in one sentence bag that contain lots of objective facts and information about the same entity pair for DSRE. 
Although previous works achieved good relation extraction results~\cite{2016BiGRU-ATT,2020bere}, they mostly adopted \revised{a} selective attention mechanism to handle \revised{the} noisy labeling problem, which mainly used information from valid sentences, and neglected potential relevance between sentences within a bag and \revised{the} loss of background information contained in noisy sentences for target entity pair. 
\revised{The} methods could not \revised{effectively} use the inter-sentence level information and the background information contained in noisy sentences. \par

\begin{table}[h]
    \centering
    \caption{An example of a biomedical distant supervision sentence bag.}
    \begin{tabular}{ccc}
        \toprule
        {ID} & {Sentence} & {Valid} \\
        \midrule
        {a} & \makecell{Calcium hopantenate, which is obtained \\by substituting the \textbf{beta-alanine} of \\pantothenic acid for gamma-\textbf{amino} butyric \\acid, is a therapeutic drug for mental \\retardation and cerebrovascular dementia.} & {False} \\
        \midrule
        {b} & \makecell{Uptake of gaba was inhibited by \\beta-Guanidinopropionic acid, \textbf{beta-alanine}, \\gamma-amino-beta-hydroxybutyric acid, \\beta-\textbf{amino}-n-butyric acid,\\3-aminopropanesulphonic acid and taurine.} & {False}\\
        \midrule
        {c} & \makecell{The presence of the characteristic \\ 4'-phosphopantetheine prosthetic group \\ was indicated by the occurrence of\\equimolar quantities of \textbf{beta-alanine} \\and taurine in \textbf{amino acid} hydrolysates.} & {True}\\
        \midrule
        \multicolumn{3}{l}{$ Entity \ Pair: beta-alanine,\ amino \  acid$} \\
        \multicolumn{3}{l}{$ Relation: has\_chemical\_structure $} \\
        \bottomrule
    \end{tabular}
    \label{tab:example_bag}
\end{table}

In the multi-instance learning framework, although the labels of noisy sentences are different from the label of the bag, the noisy sentences still express some information and objective facts about the same entity pair, which could be regarded as a sort of background information for \revised{the} target entity pair. 
Table \ref{tab:example_bag} gives an example of a biomedical distant supervision sentence bag, in which sentence $ c $ expresses the distant supervision relation label "$ has\_chemical\_structure $", while sentences $ a $ and $ b $ do not, and their tail entities are mislabeled as “$ amino $”.
Although the noise sentences $ b $ and $ c $ do not directly express relation "$ has\_chemical\_structure $"  between entities "$ beta-alanine $" and "$ amino \  acid $", they both describe the biomedical facts about beta-alanine and other kinds of amino acids, and express that there are semantic relations closely related to "$ has\_chemical\_structure $" between entities, such as "$ substitute $".
We hypothesize that there may be relevance among these sentences, and discovering and exploiting the inter-sentence-level information could help to predict the correct relation of sentence bag. 
And, in the subsequent experiments, we also demonstrate the existence of relevance between these sentences.
Hence, how to exploit the relevance over sentences within a bag and make full use of inter-sentence level information becomes \revised{a problem of interest} in biomedical relation extraction.\par

Compared with the relation extraction datasets in \revised{the} general \revised{text mining} domain~\cite{2010Modeling}, we summarize \revised{unique} characteristics of biomedical distant supervision datasets as follows: 
(1) the number of instances within a sentence bag is \revised{larger, so a} sentence bag will contain a larger amount of information, and there may be relevance between sentences.
(2) the content of \revised{a} sentence is based on objective biomedical facts, and \revised{opposed descriptions of the same entity pair in different sentences are rarely observed.} 
Therefore, although biomedical sentences contain noise, we \revised{cannot} just filter a fixed number of key words or discard noisy words, because the content of the entire sentence is informative for the task.
(3) the acquisition and processing of external biomedical information \revised{is} more difficult, \revised{requiring} lots of efforts of experts and researchers in related fields.
(4) the number of relations in general biomedical datasets is very large, which reflects the large amount of relations mentioned in biomedical publications. 
\revised{These characteristics challenge biomedical relation extraction models to further improve performance.}
\par

Based on the above motivations and characteristics of biomedical datasets, we propose a novel graph-based biomedical relation extraction framework (GBRE). 
In our framework, we first use query-sentence attention to capture the key words in sentence that are more critical to the relation between target entity pair and reduce sentence noise. 
Then, inspired \revised{by the graph} attention network (GAT)~\cite{2017Graph}, in which we view sentences and a sentence bag as nodes and a fully connected graph, respectively, and encode rich neighborhood information of the graph via an intra-bag self-attention mechanism. 
In this way, the relevance between sentences can be explored and learned, and the inter-sentence level information in the bag can be effectively utilized.
Additionally, our proposed method does not require introduction of external knowledge or construction of rules such as constraints, and can be directly applied to any biomedical distant supervised dataset.
We evaluate our network on BioRel~\cite{2019BioRel}, the largest domain-general biomedical dataset for distant supervised relation extraction, and TBGA~\cite{2022tbga}, the largest available dataset for Gene-Disease Association (GDA) extraction, and achieve the best \revised{relation extraction results.} 
\revised{
Additionally, we also evaluate our proposed GBRE approach on NYT-10~\cite{2010Modeling}, 
the public mainstream general text-mining DSRE benchmark,
where again our network outperforms the state-of-the-art baselines.
These experimental results demonstrate the excellent performance and universality of the proposed method for relation extraction.
}
Furthermore, we conduct ablation studies and present and dissect an illustrative example to demonstrate that the methods we propose are highly effective for \revised{the} biomedical relation extraction task. 
% We make the datasets and source code publicly available at https://github.com/nclabhzhang/GBRE.
\par

The contributions of our  work can be summarized as follows:
\begin{enumerate}
\item \revised{We propose a graph model for a sentence bag, and an associated intra-bag self-attention mechanism, 
which effectively capture the relevance between sentences and utilize the inter-sentence level information for the sentence bag.}
\item \revised{We develop novel query generation method and combine it with query-sentence bidirectional attention, 
to reduce word-level noise for DSRE.
The method effectively alleviates the noisy labeling problem, 
in combination with selective attention without requiring additional mechanisms for modeling external information
(e.g. constraint rules or entities descriptions).}
\item \revised{We demonstrate the excellent performance and universality of the proposed GBRE method in comparison with several state-of-the-art methods for relation extraction in biomedical and general text-mining domains, via large-scale experiments on two biomedical datasets and the NYT-10 dataset.}
\end{enumerate}

\section{Related Works}

As an important subtask of information extraction in the field of
natural language processing, relation extraction was
\revised{originally} treated as a supervised learning task\revised{,
  to which, significant research effort has been devoted}
\cite{2012socher,2015santos,2016cai,2019lee}.  Due to the large
requirement of time and effort for supervised training, Mintz et
al.~\cite{mintz2009distant} proposed distant supervision approach to
automatically generate large scale labeled training data. \par

However, \revised{the} DSRE assumption always suffers from
\revised{the} noisy labeling problem.  Hence, many works
\cite{2010Modeling,2011Knowledge,2012Multi} viewed DSRE as
multi-instance learning problem, which aims to extract relations of an
entity pair from a sentence bag instead of a single sentence, to
alleviate the noisy labeling problem.  On the basis of multi-instance
learning, many works have proposed novel noise reduction methods, such
as attention mechanisms
\cite{2015Distant,2016Neural,2017Distant,2018Improving,2019batt} and
external information integration, e.g. entity information
\cite{2021Improving,2014Exploring}, knowledge graph information
\cite{2018Label} and constraint rules~\cite{2022liang}.  Li et
al.~\cite{2019Entity} consider the entity-relation extraction task as
a multi-turn question-answering task and show that the methods of
question-answering~\cite{2021robust} and related attention mechanisms
\cite{2017bidirectional,2017coattention} \revised{are} useful for
relation extraction. They believe that \revised{the} question query
encodes important prior information for the relation class we want to
identify.  In our proposed method, GBRE automatically generates
\revised{a} generic query for each sentence without requiring text
preprocessing or the introduction of external information, and its
attention mechanism integrates the prior information of \revised{the}
query, which is effective for capturing key words in \revised{a}
sentence and reducing sentence noise. \par

In the biomedical domain, many works have been devoted to supervised
relation extraction tasks and have achieved desirable results, such as
\revised{identifying} protein–protein interactions (PPIs)~\cite{2017PPI} and drug-drug interactions (DDIs)~\cite{2014DDI,2022DDI}. As mentioned above, most research on
biomedical distant supervision relation extraction is also inspired by~\cite{mintz2009distant},
such as rule induction~\cite{2012ravikumar,2014liu}, \revised{a} variant of multi-instance
learning~\cite{2017Extracting} and neural networks, e.g. combining
with reinforcement learning strategy~\cite{2021Distantly+RL}.  To
reflect the large number of relations mentioned in biomedical
publications and the real distribution of relations, more and more
large-scale biomedical DS datasets are \revised{being} proposed
\cite{2019BioRel,2022tbga}, which \revised{contain} more instances and
information in the sentence bag. \revised{Recently, transformer 
architecture~\cite{2017attention} based models, such as BERT~\cite{2019bert},
have revolutionized NLP tasks and have also been 
applied to biomedical relation extraction~\cite{2020biobert} and
general DSRE~\cite{2022pare}.  Additionally, contrastive learning
frameworks~\cite{2021cil,2022hiclre} have also been adopted for DSRE
in combination with BERT.}
Although neural network methods~\cite{2016Neural,2016BiGRU-ATT,2020bere} achieve promising results for
these large-scale biomedical datasets, they are still far from
satisfactory.  Different from \revised{the} selectivity of
attention-based models, that mainly utilize information from valid
sentences, our proposed framework GBRE is capable of learning the
relevance between sentences and \revised{utilizing} the inter-sentence
level information of \revised{a} sentence bag and the background
information of noise sentences for entity pairs.  By viewing
\revised{a} sentence bag as graph-structured data, each sentence
aggregates information from its neighbors according to the degree of
relevance between sentences. By integrating the above information,
GBRE improves the utilization of sentence bags and effectively
alleviates the interference of noisy labels.

\section{GBRE Framework}

\begin{figure*}
    \centering
    \includegraphics[width=0.96\textwidth]{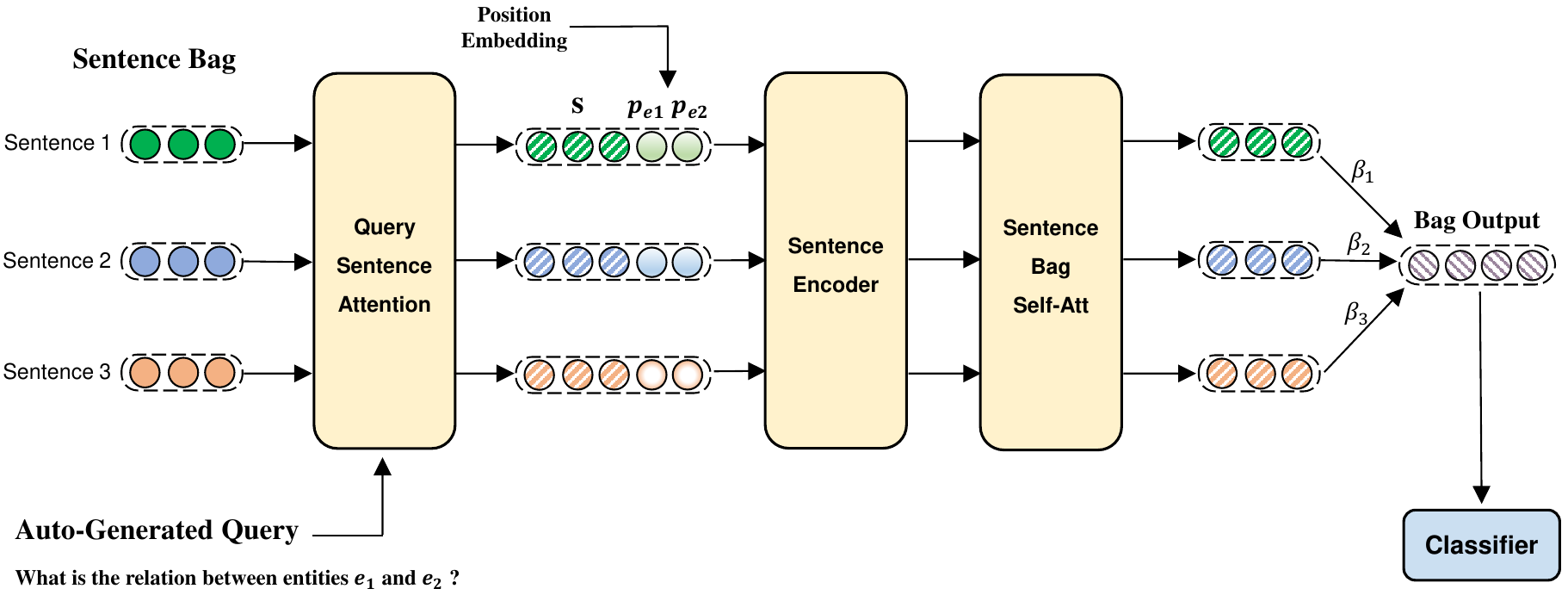}
    \caption{The proposed graph-based relation extraction framework (GBRE). Query-sentence attention is adopted to couple a query vector and sentence vector and produce a set of query-aware feature vectors for the sentence. A sentence encoder is used to obtain the sentence representations. The bag self-attention layer aims to extract the relevance between sentences within a bag and utilize the inter-sentence level information of a sentence bag by viewing the sentence bag as graph. A selective attention layer is used to obtain the sentence bag representation by performing a weighted sum on the representations of sentences. A final classifier predicts relations mentioned in the sentence bag.}
    \label{fig:model}
\end{figure*}

Given a bag of sentences
$ B = \left \{ s_{1}, s_{2},...,s_{N} \right \} $ and a corresponding
entity pair $\left ( e_{1}, e_{2} \right ) $, the objective of distant
supervision relation extraction is to predict the relation $ r $
between the two entities $\left ( e_{1}, e_{2} \right )
$. \revised{Our proposed approaches for treating the sentence bag as a
  graph and for using a synthesized query to exploit query sentence
  attention mechanisms are versatile and can be effectively
  incorporated into alternative DSRE pipelines. We demonstrate this by
  constructing and presenting results for two alternative DSRE
  pipelines. To make our mainline description self-contained, we focus
  on the more modular and simpler of these two pipelines and defer a
  description of the second alternative to the Appendix, which relies
  on cited references for details. The DSRE pipeline that is the focus
  of our mainline description is illustrated in Fig.~\ref{fig:model},
  organized as three main modules:}
\begin{itemize}

\item \textbf{Query-Sentence Attention.} Given a sentence, query-sentence attention is adopted to couple the information of query vector and sentence vector, and produce a set of query-aware feature vectors for the sentence.

\item \textbf{Sentence Encoder.} Given a sentence vector, a sentence encoder is used to represent it as a \revised{reduced dimensionality} vector.

\item \textbf{Sentence Bag Self-Attention.} Given the representations of all sentences within \revised{a} bag $ B $, sentence bag self-attention aims to derive the relevance over sentences within the bag via message passing based on the graph structure.

\end{itemize}
Details of the architecture are presented in the following subsections.

\subsection{Input Word Vector Mapping Layer} \label{sec:WordEmbeddingLayer}

The input layer aims to map sentence and query words into \revised{a vector representation} that captures semantic and syntactic information.
Given a sentence $ s $ that mentions a head-tail entity pair $\left ( e_{1}, e_{2} \right ) $, 
we first generate a general query "What is the relation between head-entity $e_{1}$ and tail-entity $e_{2}$?" for the sentence. 
This automatically generated query encodes important prior information for the semantic relation expressed by the entity pair that we want to identify. 
The word vector embedding transforms each word $ w_{l} $ in the sentence $ s $ and each word $ q_{t} $ in the query $ q $ into $ d_{w} $-dimensional vectors $ \textbf{w}_{l}, 1 \le l \le L$ and $ \textbf{q}_{t}, 1 \le t \le T $ respectively via a pre-trained word embedding matrix. Thus, we obtain the vector representations of \revised{a} sentence and \revised{the} corresponding query, and denote them as vector sequences $ S = \left \{\textbf{w}_{1},\textbf{w}_{2},...,\textbf{w}_{L} \right \} $ and $ Q = \left \{\textbf{q}_{1},\textbf{q}_{2},...,\textbf{q}_{T} \right \} $, where $ \textbf{w}_{l}, \textbf{q}_{t} \in \mathbb{R}^{d_{w} } $.

\subsection{Query-Sentence Attention}

Query-sentence attention aims to merge the information of sentence words and query words, and capture the key words in a sentence that are more critical to the relation between the target entity pair, which can reduce the sentence noise.
Our method outputs \revised{a} new representation of the whole sentence with original length, which has interacted with the query statement and integrated important information about the relation we want to identify.
The representation can be fed into \revised{an} encoder to further extract high-dimensional features, 
or can be directly sent to \revised{a} classifier layer for relation prediction.
In order to better integrate the information of \revised{a} sentence and query, we follow~\cite{2017bidirectional} and calculate attention scores in two directions: from sentence to query and from query to sentence.\par

Given an input sentence sequence $ S = \left \{\textbf{w}_{1},\textbf{w}_{2},...,\textbf{w}_{L} \right \} $ and a corresponding input query sequence $ Q = \left \{\textbf{q}_{1},\textbf{q}_{2},...,\textbf{q}_{T} \right \} $, we first calculate the similarity matrix $ H\in \mathbb{R}^{L \times T}  $, between embedding sequences $ S $ and $ Q $ as
\begin{equation}
\begin{split}
    H_{lt} &=  \phi \left ( \textbf{w}_{l},\textbf{q}_{t} \right ) \\
    \revised{\phi} ( \textbf{w},\textbf{q} ) &= \textbf{W}_{h} \left [ \textbf{w} ; \textbf{q} ; \textbf{w} \circ \textbf{q} \right ] \\
\end{split}
\end{equation}
where $ H_{lt}\in \mathbb{R} $ denotes the similarity between the $ l $-th sentence word and the $ t $-th query word, 
$ \textbf{w}_{l} $ is the $ l $-th word vector of sentence representation $ S $, $ \textbf{q}_{t} $ is the $ t $-th word vector of query representation $ Q $, 
$ \phi(\textbf{w},\textbf{q})\in \mathbb{R} $ is a trainable scalar function that calculates the similarity score between input vectors $ \textbf{w} $ and $ \textbf{q} $ \revised{via a trainable weight vector $ \textbf{W}_{h}\in \mathbb{R}^{3d_{w}} $}, and $ \circ $ denotes element-wise multiplication.\par

\begin{figure}
    \centering
    \includegraphics[width=0.4\textwidth]{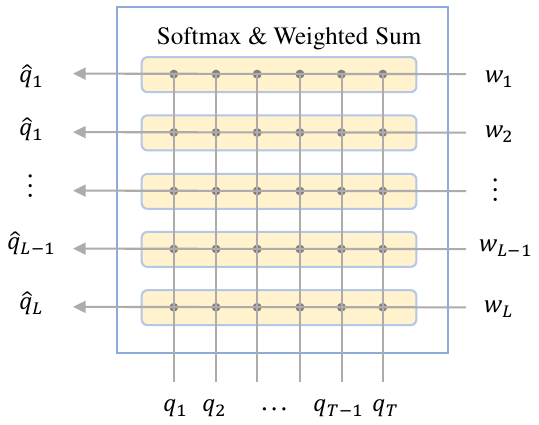}
    \caption{Sentence to query ($sq$) attention score computation.}
    \label{fig:s2q}
\end{figure}
\begin{figure}
    \centering
    \includegraphics[width=0.4\textwidth]{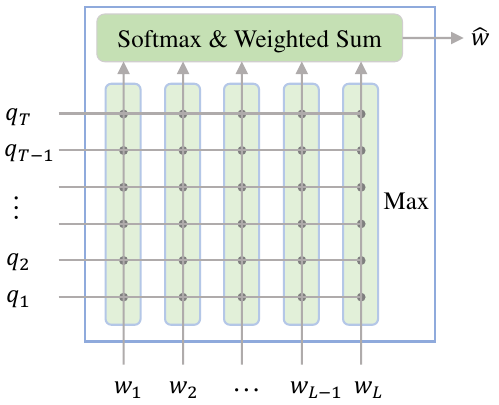}
    \caption{Query to sentence ($qs$) attention score computation.}
    \label{fig:q2s}
\end{figure}

Then, as shown in Figs.~\ref{fig:s2q} and~\ref{fig:q2s}, we calculate the attention scores and feature vectors in both directions, from sentence to query (sq) and the reverse query to sentence (qs) direction. The sentence to query (sq) attention scores are computed  as 
\begin{equation}
    \alpha_{l}^{(\textrm{sq})} = \mbox{Softmax}(H_{l:} )
\end{equation}
\begin{equation}
    \hat{\textbf{q}}_{l} = \sum_{t=1}^{T}\alpha_{lt} ^{(\textrm{sq})}\textbf{q}_{t}   \\
\end{equation}
where $ \alpha _{l}^{(\textrm{sq})}\in \mathbb{R}^{T}  $ denotes the attention scores for query words to the $ l $-th sentence word, $ \hat{Q} = \left \{ \hat{\textbf{q}}_{1},\hat{\textbf{q}}_{2},...,\hat{\textbf{q}}_{L}  \right \} \in \mathbb{R}^{L\times d_{w}} $ is the sentence representation combined with query information, and $ {\textstyle \sum_{t=1}^{L}} \alpha _{lt}^{(\textrm{sq})} =1 $ for all $ l $. The  query to sentence (qs) attention scores are computed as
\begin{equation}
    \alpha^{(\textrm{qs})} = \mbox{Softmax}(\max_{\textrm{col}}(H)) 
\end{equation}
\begin{equation}
    \hat{\textbf{w}}  = \sum_{l=1}^{L} \alpha_{l}^{(\textrm{qs})} \textbf{w}_{l} 
\end{equation}
where $ \alpha^{(\textrm{qs})}\in \mathbb{R}^{L} $ denotes attention weights on the sentence words, and $ \hat{\textbf{w}} $ denotes the weighted sum of the most important word in the sentence for the query. $ \hat{\textbf{w}} $ is tiled $ L $ times and giving $ \hat{S} \in \mathbb{R}^{L\times d_{w} } $.\par

Finally, we couple $ \hat{S} $ and $ \hat{Q} $ to generate $ \textbf{S} $, the new representation of the sentence:
\begin{equation}
    \hat{\textbf{x}}_{l} =  \varphi(\textbf{w}_{l},\hat{\textbf{q}} _{l},\hat{\textbf{w}} )
\end{equation}
\begin{equation}
    \textbf{S} = \left \{ \hat{\textbf{x}}_{1},\hat{\textbf{x}}_{2},...,\hat{\textbf{x}}_{L}  \right \}
\end{equation}
where $ \hat{\textbf{x}}_{l} \in \mathbb{R}^{3d_{w}} $, $ \varphi(\textbf{w},\hat{\textbf{q}},\hat{\textbf{w}}) = \left [\textbf{w};\textbf{w} \circ \hat{\textbf{q}};\textbf{w} \circ \hat{\textbf{w}}\right ] $, and $ \circ $ is element wise multiplication.

\subsection{Sentence Encoder} \label{sec:SentenceEnc}
The sentence encoder aims to extract a \revised{reduced dimensionality} representation from the input vector sequence. 
In order to describe the position information of a target entity pair $\left ( e_{1}, e_{2} \right ) $ in a sentence, we adopt position features~\cite{2014Relation} in our work.
Vectors $ \textbf{p}_{l}^{e_{1}} $ and $ \textbf{p}_{l}^{e_{2}} $ of $ d_{p} $-dimension are used to embed the relative distances between word $ w_{l} $ and target entities. \par

Given the input vector sequence of a sentence $ \textbf{S} = \left \{ \hat{\textbf{x}}_{1},\hat{\textbf{x}}_{2},...,\hat{\textbf{x}}_{L}  \right \} $, we concatenate it with its position embedding vectors $ \textbf{p}_{l}^{e_{1}} $ and $ \textbf{p}_{l}^{e_{2}} $ to incorporate the position information as follows:

\begin{equation}
    \textbf{x}_{l} = \left[\hat{\textbf{x}}_{l};\textbf{p}_{l}^{e_{1}};\textbf{p}_{l}^{e_{2}}\right ] \in \mathbb{R} ^{3d_{w}+2d_{p}}, 1\le l \le L
\end{equation}
\begin{equation}
    \bar{\textbf{S}} = \left \{ {\textbf{x}_{1}},{\textbf{x}_{2}},...,{\textbf{x}_{L}}\right \}
\end{equation}

The sentence encoder PCNN slides the convolutional kernels $ {K}_{c} $ over $ \bar{\textbf{S}} $ to capture the hidden representations as follows:
\begin{equation}
    \textbf{m}_{i} = \revised{K_{c}}\textbf{x} _{i-w+1:i}\in \mathbb{R}^{L},1 \le i \le c
\end{equation}
where $ \textbf{x} _{m:n} $ is $ \left [\textbf{x} _{m},\textbf{x} _{m+1},...\textbf{x} _{n}\right ] $, 
and $ c $ \revised{indexes over the kernels.}\par

Then, piece-wise max pooling is used to extract features from the three segments of convolution outputs:
\begin{equation}
    \begin{split}
        & \textbf{u}_{i}^{(1)} = \max_{1\le j \le k_{1}} \left ( \textbf{m}_{ij}\right ) \\
        & \textbf{u}_{i}^{(2)} = \max_{k_{1}\le j \le k_{2}} \left ( \textbf{m}_{ij}\right ) \\
        & \textbf{u}_{i}^{(3)} = \max_{k_{2}\le j \le L} \left ( \textbf{m}_{ij}\right ) \\
    \end{split}
\end{equation}
where $ k_{1} $ and $ k_{2} $ are \revised{the} positions of target entities $ e_{1} $ and $ e_{2} $ in the sentence. Then, we can obtain the piece-wise max pooling result $ u_{i} = {\left \{ u_{i}^{(1)}, u_{i}^{(2)},u_{i}^{(3)} \right \} }  $ of the $ i $-th convolutional kernel.
Finally, by concatenating the pooling results and an activation nonlinearity, we obtain the sentence representation \textbf{s} as follows:
\begin{equation}
    \textbf{s} = \sigma(\textbf{u}_{1:c})\in \mathbb{R}^{3c} 
\end{equation}
where $ \sigma (\cdot) $ is the activation function, a rectified linear unit (RELU) in our implementation.

\subsection{Sentence Bag Self-Attention} \label{sec:SentBagGraphAttention}
\begin{figure}
    \centering
    \includegraphics[width=0.45\textwidth]{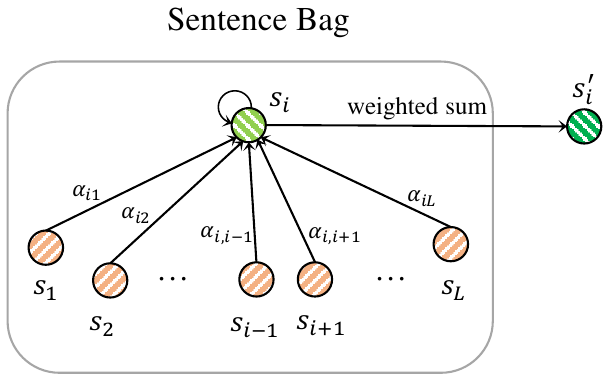}
    \caption{Sentence bag graph structure. Each node $ s_{i} $ denotes the corresponding sentence and the sentence bag is viewed as graph.}
    \label{fig:sentence_bag_structure}
\end{figure}

Sentence bag self-attention is a graph based attention mechanism that aims to derive the relevance between sentences within a bag. 
Inspired by GAT~\cite{2017Graph}, we propose sentence bag self-attention to convert the sentence bag into a graph structure and then encode the information of the whole sentence bag. 
In this layer, each node gathers and summarizes information from  all its immediate neighbors; thus, information is conveyed along the edges of the graph. 
The attention mechanism can encode rich neighborhood information of the graph, as shown in Figure \ref{fig:sentence_bag_structure}.\par

Given a bag of sentences representations $ B=\left \{\textbf{s}_{1},\textbf{s}_{2},...,\textbf{s}_{L} \right \} $, we regard each sentence in the bag as a node in an undirected fully connected graph, where the edge strengths between nodes are calculated by the attention mechanism. First, self-attention is used on the sentences in bag to calculate attention coefficients as follows:
\begin{equation}
    e_{ij} = \mbox{Similarity}\left ( \textbf{s}_{i},\textbf{s}_{j} \right )
\end{equation}
where $ \mbox{Similarity}\left ( \textbf{s}_{i},\textbf{s}_{j} \right ) $ is a function to calculate the similarity between two input sentences, we adopt cosine similarity as the function\revised{,  and }$ e_{ij} $ denotes the importance of sentence $ \textbf{s}_{j} $'s features to the sentence $ \textbf{s}_{i} $. 
Over all choices of the sentence $ j $, we normalize these similarities using the softmax function to obtain values $ \alpha _{ij} $  that indicate the degree of relevance between sentences $ \textbf{s}_{i} $ and $ \textbf{s}_{j} $, i.e,
\begin{equation}
    \alpha _{ij} = \frac{\exp (e_{ij} ) }{ {\textstyle \sum_{k}^{}\exp (e_{ik})} } 
    \label{eq:bag_att_score}
\end{equation}
Finally, an updated representation of sentence $ \textbf{s}_{i} $ is obtained as
\begin{equation}
    \textbf{s}'_{i} =  \sum_{j=1}^{L} \alpha_{ij} \textbf{s}_{j}
\end{equation}

The new representation $ \textbf{s}'_{i} $ of sentence $ \textbf{s}_{i} $ integrates relevant information of the neighborhood sentences. In other words, the sentence node in the graph has fused rich information from  all its immediate neighbor nodes in graph. This process yields a vector representation for each sentence in a bag.

\subsection{Selective Attention Layer}
Following previous work~\cite{2016Neural}, we use selective attention over instances to obtain the sentence bag representation. Given a bag of sentences, the attention weight $ \beta _{i}  $ of the $ i $-th sentence for its corresponding relation $ r $ is calculated as follows:
\begin{equation}
    \begin{split}
        & {c}_{i} = \text{s}'_{i}\textbf{A} \textbf{r} \\
        & {\beta }_{i} = \frac{\exp({c}_{i})}{ {\textstyle \sum_{k}^{}\exp({c}_{k})} } \\
    \end{split}
\end{equation}
where \textbf{r} is a trainable vector which denotes the representation of the relation $ r $, \textbf{A} is a weighted diagonal matrix, and $ {c}_{i} $ is a query-based function that scores how well the input sentence $ \text{s}_{i} $ matches the predicted relation $ r $. 
The bag representation is then derived as the weighted sum of sentence representations:
\begin{equation}
    \textbf{z} =  {\sum_{i=1}^{N} \beta_{i} \textbf{s}'_{i} }
\end{equation}
where $ N $ is the number of the sentences in the sentence bag $ B $.\par

Finally, the bag representation \textbf{z} is fed into a softmax classifier to compute a probability distribution over relation labels as follows:
\begin{equation}
    P(r|B;\Theta )=\mbox{Softmax}(\textbf{W}\textbf{z}+\textbf{b} )
\end{equation}
where $ \Theta $ is the set of model parameters, and \textbf{W} and \textbf{b} represent the classifier weights and bias, respectively.

\subsection{Optimization}
The bag level objective function is defined as the cross-entropy loss
\begin{equation}
    J(\Theta )=-\frac{1}{M}\sum_{i=1}^{M}\log P(r_{i}|\textbf{z}_{i};\Theta )
\end{equation}
where $ M $ is the number of sentence bags, $ r_{i} $ is the relation label of bag $ B_{i} $, $ \textbf{z}_{i} $ is the representation of bag $ B_{i} $, and $ \Theta $ represents all the parameters of the model. 
The model parameters are estimated by minimizing the objective function $ J(\Theta) $ through mini-batch stochastic gradient descent (SGD)~\cite{1951sgd,2019sgdtorch}.

\section{Experimental Results and Discussion}

We conducted comprehensive experiments to evaluate the performance of the proposed method.
First, we introduce the benchmark datasets for biomedical distant supervised relation extraction and evaluation metrics used in the experiments. Then, we describe the hyper-parameters settings of our experiments. Finally, we compare the performance of our method with several competitive baseline methods, conduct ablation experiments to highlight the contribution of the individual components in our framework, and present and dissect an illustrative example to demonstrate the effectiveness of our proposed method.

\subsection{Datasets and Evaluation Metrics}

\begin{table}[t]
        \centering
        \caption{Statistics of BioRel and TBGA datasets.}
        \begin{tabular}{cccccc}
        \toprule  % 顶部线
             {\textbf{Dataset}} & {\textbf{Split}} & {\textbf{Instances}} & {\textbf{Bags}} & {\textbf{Ins.s/bag}} & {\textbf{Relations}}\\
        \midrule
             \multirow{3}{*}{BioRel} & {Train} & {534,277} & {39,969} & {13.37} & \multirow{3}*{125} \\
             & {Valid} & {114,506} & {20,675} & {5.54} \\
             & {Test}  & {114,565} & {20,756} & {5.52} \\
        \midrule
             \multirow{3}{*}{TBGA} & {Train} & {178,264} & {85,047} & {2.10} & \multirow{3}*{4} \\
             & {Valid} & {20,193} & {10,491} & {1.92} \\
             & {Test}  & {20,516} & {10,494} & {1.94} \\
        \bottomrule
        \end{tabular}
        \label{tab:dataset_statistics}
    \end{table}

Two biomedical benchmark datasets are adopted in our experiments:
\begin{itemize}[leftmargin=*]
\item \textbf{BioRel}\cite{2019BioRel}, a large-scale domain-general biomedical dataset for distant supervision relation extraction, is constructed by aligning the knowledge base UMLS~\cite{2004umls} with the corpus source MEDLINE. It consists of 124 labels corresponding to actual relations and a NA (not a relation) label, and contains more than 530,000 sentences. BioRel has less noisy data and is suitable for relation extraction using deep learning methods.
\item \textbf{TBGA}\cite{2022tbga}, the largest available dataset for GDA extraction, is generated by using DisGeNET~\cite{2019DisGeNET} as a source database and several expert-curated resources. TBGA reflects the sparseness of GDAs in biomedical literature and is a challenging dataset for automatic GDA extraction, one of the most relevant \revised{biomedical relation extraction tasks}.
\end{itemize}
Table \ref{tab:dataset_statistics} shows the overall statistics for BioRel\footnotemark[1] and TBGA\footnotemark[2]. \par
\footnotetext[1]{https://bit.ly/biorel\_dataset}
\footnotetext[2]{https://zenodo.org/record/5911097}

Following previous studies~\cite{2019BioRel} and~\cite{2022tbga}, precision-recall (PR) curves, area under curve (AUC) values and Precision@N (P@N) values~\cite{2016Neural} are adopted as evaluation metrics in our experiments.

\subsection{Hyper-Parameter Settings}
\begin{table}[t]
        \centering
        \caption{Hyper-parameter settings for the models for BioRel and TBGA.}
        \begin{tabular}{cccc}
        \toprule  % 顶部线
             \multirow{2}*{Component} & \multirow{2}*{Parameters} & \multicolumn{2}{c}{Value} \\
             \cmidrule(lr){3-4}
             & { } & {BioRel} & {TBGA}  \\
        \midrule
             \multirow{2}{*}{\makecell{Query-Sentence\\Attention}} & {\revised{word size}} & {200} & {200}  \\
             & {output size} & {600} & {600}  \\
        \midrule
             \multirow{4}{*}{\makecell{Sentence\\Encoder}} & {\revised{hidden size}} & {230} & {230}  \\
             & {\revised{output size}}  & {690} & {690}  \\
             & {window size} & {3} & {3}  \\
             % & {word size}  & {200} & {200}  \\
             & {position size}  & {5} & {5}  \\
        \midrule
            {\makecell{Sentence Bag\\Self-Attention}} & {dropout rate} & {0.3} & {0.25}  \\
        \midrule
            {Classifier} & {input size} & {690} & {690}  \\
        \midrule
             \multirow{4}{*}{Optimization} & {learning rate} & {0.05} & {0.1}  \\
             & {dropout rate} & {0.5} & {0.5}  \\
             & {batch size}  & {30} & {128}  \\
             & {\revised{optimizer}}  & {SGD} & {SGD}  \\
        \bottomrule
        \end{tabular}
        \label{tab:parameters}
    \end{table}
All of the hyper-parameters used in our experiments are listed in Table \ref{tab:parameters} \revised{for the set-up using the PCNN sentence encoder. Corresponding values for the set-up with the BERT-based sentence encoder are provided in the Appendix in Table~\ref{tab:bert_parameters}.}
For a fair comparison, we set most of the hyper-parameters identical to~\cite{2019BioRel} and~\cite{2022tbga}. The 200-dimensional word embeddings released by these prior works are also adopted for initialization. 
All weight matrices and position embeddings are initialized by Xavier initialization~\cite{glorot2010understanding}, and the bias vectors are all initialized to 0. 
A batch of sentence bags are randomly selected from the training set and fed to proposed model for each iteration until convergence.

\subsection{Performance Comparison}

\begin{figure*}[ht]
        \centering
        \scriptsize
        \begin{tabular}{cc}
             \includegraphics[width=0.47\textwidth]{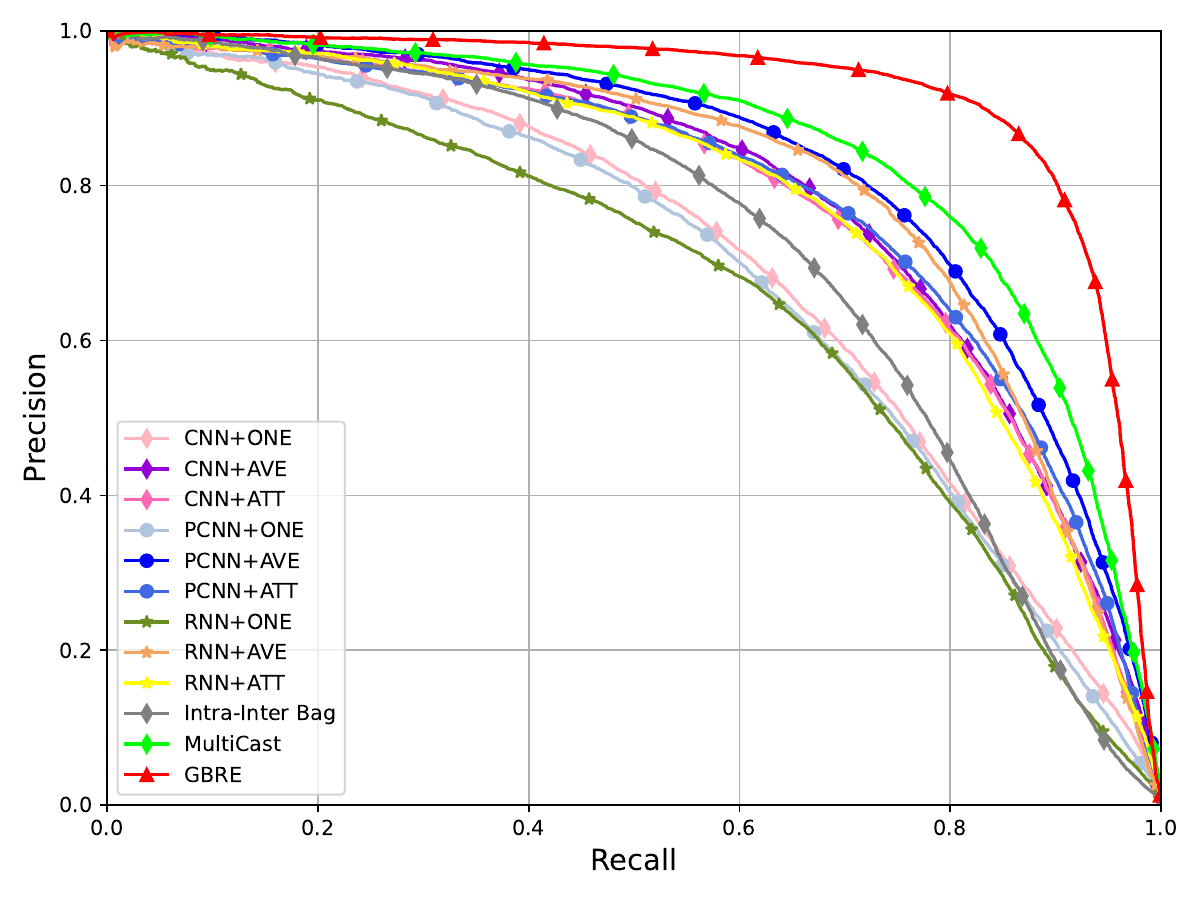} & 
             \includegraphics[width=0.47\textwidth]{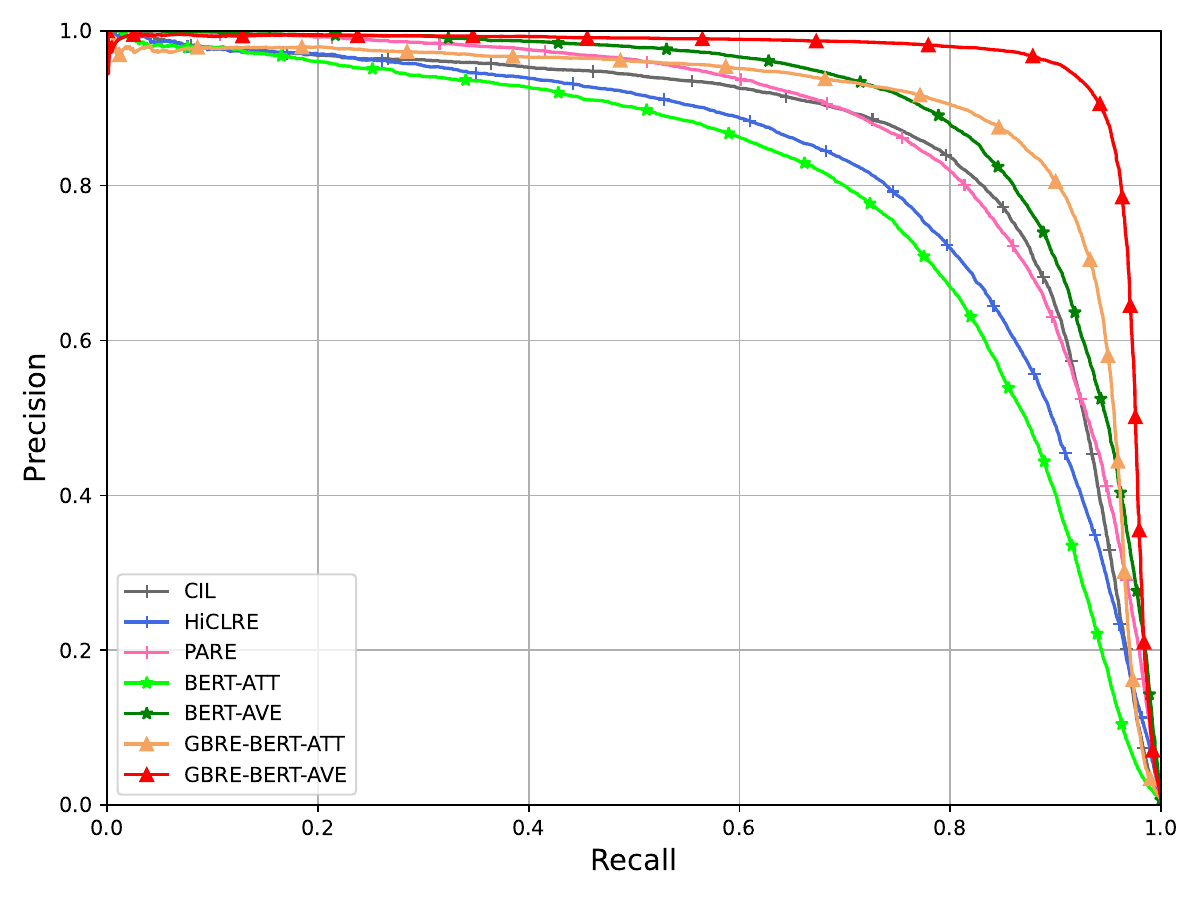} \\
             (a) & (b) \\
             \includegraphics[width=0.47\textwidth]{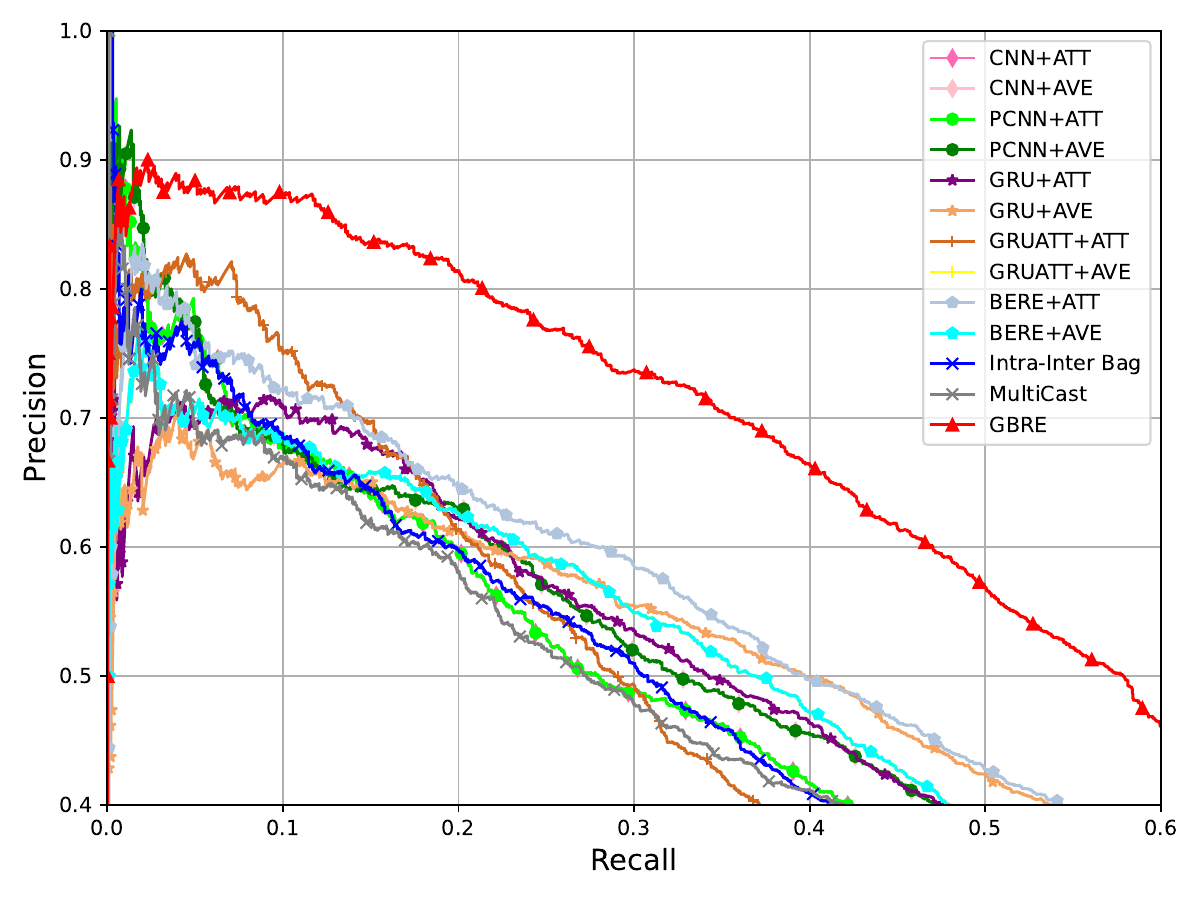} & 
             \includegraphics[width=0.47\textwidth]{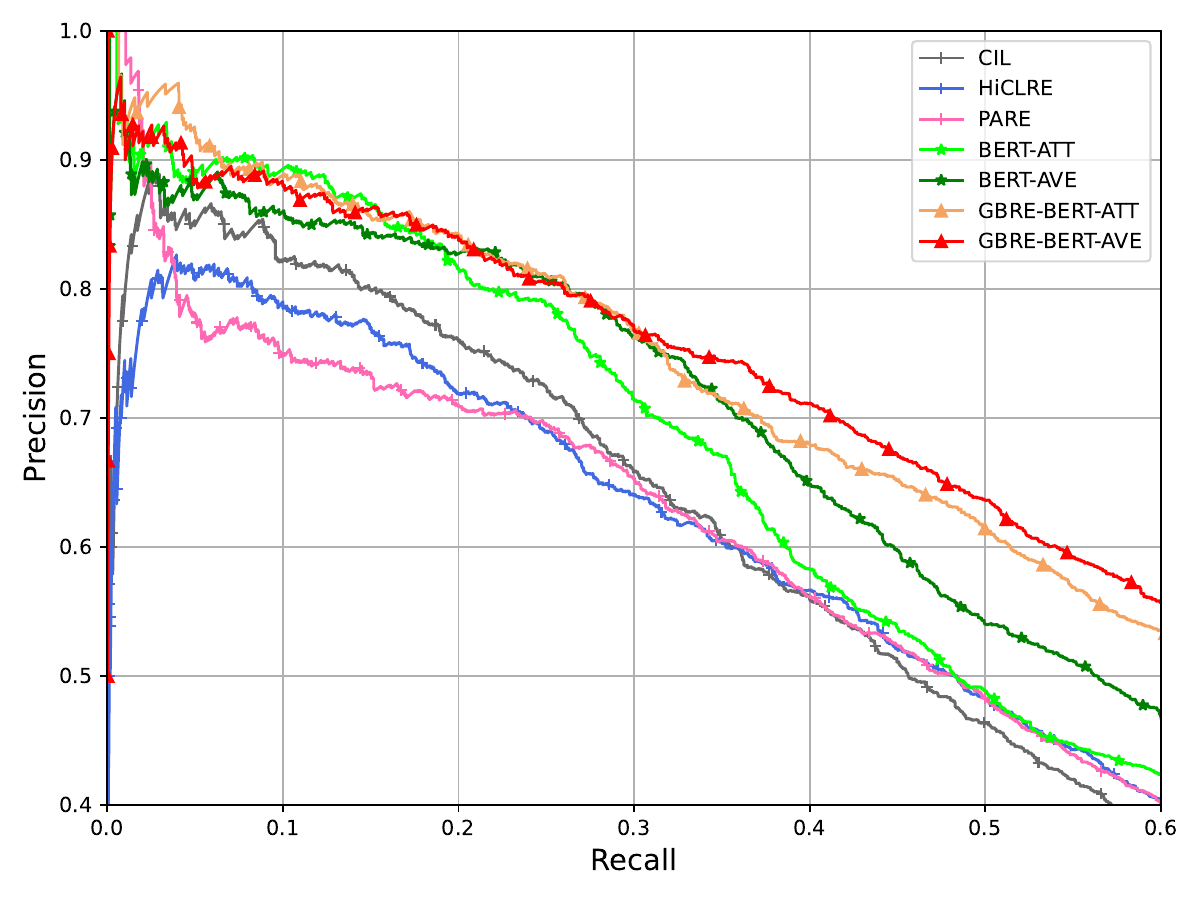} \\
             (c) & (d) \\
        \end{tabular}
        \caption{PR curves over the BioRel and TBGA datasets for the proposed GBRE model and for several prior methods. The proposed GBRE model exhibits the best performance on both datasets.
        \revised{
        Note that GBRE-BERT indicates BERT-based GBRE variant, the proposed GBRE model using BERT as encoder layer.
        (a) non-BERT models on BioRel dataset. (b) BERT-based models on BioRel dataset.
        (c) non-BERT models on TBGA dataset. (d)BERT-based models on TBGA dataset.
        }
        }
        \label{fig:PR_compare}
\end{figure*}

\begin{table*}[ht]
        \centering
        \caption{Performance metrics (\%) for the different non-BERT RE models on the BioRel dataset.}
        \begin{tabular}{ccccccccc}
        \toprule
             \multicolumn{2}{c}{\textbf{Model}} & \textbf{P@4000} & \textbf{P@8000} & \textbf{P@12000} & \textbf{P@16000} & \textbf{Mean} & \textbf{F1} & \textbf{AUC}\ ($\uparrow$) \\
        \midrule
            \multirow{3}{*}{CNN} & ONE & 93.4 & 84.9 & 75.0 & 65.7 & 79.8 & 66 & 70 \\
             & AVE & 94.0 & 91.0 & 81.6 & 72.0 & 85.3 & 72 & 79 \\
             & ATT & 96.4 & 90.6 & 82.3 & 72.3 & 85.4 & 72 & 78 \\
        \midrule
            \multirow{3}{*}{PCNN} & ONE & 92.1 & 83.8 & 74.5 & 65.5 & 79.0 & 65 & 70 \\
             & AVE & 96.6 & 93.6 & 85.7 & 75.4 & 88.1 & 76 & 82 \\
             & ATT & 96.2 & 91.1 & 83.3 & 73.4 & 86.0 & 73 & 79 \\
        \midrule
             \multirow{3}{*}{RNN} & ONE & 88.9 & 81.1 & 71.6 & 63.1 & 76.2 & 63 & 70 \\
             & AVE & 96.7 & 92.8 & 84.0 & 73.5 & 87.0 & 74 & 80 \\
             & ATT & 94.6 & 89.6 & 81.8 & 72.5 & 84.7 & 72 & 78 \\
        \midrule
            \multicolumn{2}{c}{Intra-Inter bag} & 95.5 & 89.1 & 78.5 & 68.0 & 82.8 & 71 & 72.4 \\
            \multicolumn{2}{c}{MultiCast} & 98.0 & 94.5 & 88.3 & 78.5 & 89.8 & 79 & 85.2 \\
        \midrule
            \multicolumn{2}{c}{GBRE} & \textbf{99.0} & \textbf{97.8} & \textbf{94.9} & \textbf{86.5} & \textbf{94.6} & \textbf{86} & \textbf{92.0} \\
        \bottomrule
        \end{tabular}
        \label{tab:AUC_compare_biorel}
    \end{table*}

\begin{table*}[ht]
        \centering
        \caption{
        \revised{Performance metrics (\%) for the different BERT-based RE models on the BioRel dataset. GBRE-BERT indicates BERT-based GBRE variant.}
        }
        \begin{tabular}{ccccccccc}
        \toprule
             \multicolumn{2}{c}{\textbf{Model}} & \textbf{P@4000} & \textbf{P@8000} & \textbf{P@12000} & \textbf{P@16000} & \textbf{Mean} & \textbf{F1} & \textbf{AUC}\ ($\uparrow$) \\
        \midrule
            \multicolumn{2}{c}{CIL} & 96.4 & 94.7 & 90.5 & 81.6 & 90.8 & 82 & 86.4 \\
             \multicolumn{2}{c}{PARE} & 98.9 & 96.6 & 90.7 & 80.7 & 91.7 & 81 & 87.7 \\
             \multicolumn{2}{c}{HiCLRE} & 96.2 & 92.6 & 86.2 & 75.6 & 87.9 & 77 & 82.4 \\
        \midrule
             \multirow{2}{*}{BERT} & AVE & 98.7 & 98.2 & 96.9 & 90.0 & 95.9 & 90 & 93.6 \\
             & ATT & 95.2 & 91.1 & 84.4 & 74.8 & 86.4 & 78 & 79.4 \\ 
        \midrule
             \multirow{2}{*}{GBRE-BERT} & AVE & \textbf{99.4} & \textbf{99.1} & \textbf{98.4} & \textbf{92.8} & \textbf{97.}4 & \textbf{93} & \textbf{96.0} \\
             & ATT & 97.5 & 96.3 & 93.4 & 86.1 & 93.3 & 88 & 90.2 \\
        \bottomrule
        \end{tabular}
        \label{tab:AUC_compare_biorel_bert}
    \end{table*}
    
    \begin{table*}[ht]
        \centering
        \caption{Performance metrics (\%) for the different non-BERT RE models on the TBGA dataset.}
        \begin{tabular}{ccccccccc}
        \toprule
             \multicolumn{2}{c}{\textbf{Model}} & \textbf{P@50} & \textbf{P@100} & \textbf{P@250} & \textbf{P@500} & \textbf{P@1000} & \textbf{Mean} & \textbf{AUC}\ ($\uparrow$)  \\
        \midrule
            \multirow{2}{*}{CNN} & AVE & 78.0 & 76.0 & 74.4 & 69.6 & 62.5 & 72.1 & 42.2\\
             & ATT & 78.0 & 76.0 & 78.8 & 71.0 & 62.4 & 73.2 & 40.3 \\
            \multirow{2}{*}{PCNN} & AVE & 78.0 & 78.0 & 74.4 & 72.0 & 66.4 & 73.8 & 42.6 \\
             & ATT & 76.0 & 75.0 & 74.4 & 70.0 & 62.8 & 71.6 & 40.4 \\
        \midrule     
            \multirow{2}{*}{BiGRU-ATT} & AVE & 74.0 & 74.0 & 74.8 & 69.4 & 61.5 & 70.7 & 41.9 \\
             & ATT & 68.0 & 76.0 & 75.6 & 70.2 & 63.1 & 70.6 & 39.0 \\
            \multirow{2}{*}{BiGRU} & AVE & 62.0 & 72.0 & 72.4 & 73.0 & 67.8 & 69.4 & 43.7 \\
            & ATT & 76.0 & 75.0 & 74.8 & 72.6 & 66.6 & 73.0 & 40.2 \\
        \midrule
            \multirow{2}{*}{BERE} & AVE & 70.0 & 71.0 & 72.0 & 70.4 & 62.0 & 69.1 & 41.9 \\
             & ATT & 78.0 & 78.0 & 80.0 & 76.4 & 70.9 & 76.7 & 44.5 \\
        \midrule
            \multicolumn{2}{c}{Intra-Inter Bag} & 76.0 & 78.0 & 74.8 & 67.6 & 61.7 & 71.6 & 40.6 \\
            \multicolumn{2}{c}{MultiCast} & 78.0 & 73.0 & 69.2 & 67.0 & 60.2 & 69.5 & 39.5\\
        \midrule
            \multicolumn{2}{c}{GBRE} & \textbf{86.0} & \textbf{89.0} & \textbf{86.8} & \textbf{86.2} & \textbf{78.7} & \textbf{85.3} & \textbf{55.3} \\
        \bottomrule
        \end{tabular}
        \label{tab:AUC_compare_TBGA}
    \end{table*}

    \begin{table*}[ht]
        \centering
        \caption{ 
        \revised{Performance metrics (\%) for the different BERT-based RE models on the TBGA dataset. GBRE-BERT indicates BERT-based GBRE variant.}
        }
        \begin{tabular}{ccccccccc}
        \toprule
             \multicolumn{2}{c}{\textbf{Model}} & \textbf{P@50} & \textbf{P@100} & \textbf{P@250} & \textbf{P@500} & \textbf{P@1000} & \textbf{Mean} & \textbf{AUC}\ ($\uparrow$)  \\
        \midrule
            \multicolumn{2}{c}{CIL} & 82.0 & 88.0 & 86.0 & 82.0 & 75.1 & 82.6 & 48.5 \\
             \multicolumn{2}{c}{PARE} & 94.0 & 88.0 & 76.4 & 74.6 & 70.7 & 80.7 & 49.4\\
             \multicolumn{2}{c}{HiCLRE} & 72.0 & 79.0 & 81.6 & 78.2 & 71.9 & 76.5 & 48.3\\
        \midrule
             \multirow{2}{*}{BERT} &  AVE & 90.0 & 89.0 & 88.8 & 85.0 & \textbf{81.9} & 86.9 & 55.7\\
             & ATT & 90.0 & 92.0 & \textbf{90.0} & \textbf{88.2} & 79.7 & 87.9 & 53.3\\ 
        \midrule
             \multirow{2}{*}{GBRE-BERT} & AVE & 92.0 & 92.0 & 88.8 & 87.0 & 81.0 & 88.2 & \textbf{59.5} \\
             & ATT & \textbf{94.0} & \textbf{95.0} & \textbf{90.0} & 87.4 & \textbf{81.9} & \textbf{89.7} & 58.3 \\
        \bottomrule
        \end{tabular}
        \label{tab:AUC_compare_TBGA_bert}
    \end{table*}

To evaluate the effectiveness of our method, 
\revised{we compare the proposed GBRE model with several competitive models and state-of-the-art models:} \par

\revised{
\begin{itemize}[leftmargin=*]
\item \textbf{CNN}~\cite{2014Relation}: a CNN-based model with only-one, average or selective attention over sentences in a bag;
\item \textbf{PCNN}~\cite{2016Neural}: a piecewise CNN-based model with only-one, average or selective attention over sentences in a bag;
\item \textbf{RNN}~\cite{2014GRU,2014GRU2}: a RNN-based model with only-one, average or selective attention over sentences in a bag;
\item \textbf{BiGRU}~\cite{2014bigru}: a bidirectional GRU-based model with average or selective attention over sentences in a bag;
\item \textbf{BiGRU-ATT}~\cite{2016BiGRU-ATT}: a BiGRU-based model with an attention layer to merge word-level features into a sentence-level feature vector;
\item \textbf{BERE}~\cite{2020bere}: a hybrid encoding network with average or selective attention over sentences in a bag;
\item \textbf{Intra–Inter Bag}~\cite{2019batt}: a PCNN-based model with intra-bag and inter-bag attention;
\item \textbf{MultiCast}~\cite{2021multicast}: a PCNN-based model integrating collaborative adversarial training, a mechanism to improve utilization of information in a sentence bag;
\item \textbf{BioBERT}~\cite{2020biobert}: a BERT-based biomedical DSRE model with average or selective attention over sentences in a bag;
\item \textbf{CIL}~\cite{2021cil}:  a BERT-based contrastive instance learning framework for DSRE;
\item \textbf{HiCLRE}~\cite{2022hiclre}: a BERT-based contrastive instance learning model integrating global structural information and local fine-grained interaction;
\item \textbf{PARE}~\cite{2022pare}: a BERT-based DSRE model in which all sentences of a bag are concatenated into a passage of sentences.
\end{itemize}
}

\revised{
Note that the first six models (i.e. CNN, PCNN, RNN, BiGRU, BiGRU-ATT and BERE) were originally applied on BioRel and TBGA, respectively.
The results for these models on the datasets are obtained from the corresponding original publications mentioned earlier.
Other results were obtained using the official source codes (i.e. Intra–Inter Bag\footnotemark[3], MultiCast\footnotemark[4], BioBERT\footnotemark[5], CIL\footnotemark[6], HiCLRE\footnotemark[7], PARE\footnotemark[8]).
}

\footnotetext[3]{https://github.com/ZhixiuYe/Intra-Bag-and-Inter-Bag-Attentions}
\footnotetext[4]{https://github.com/antct/multicast}
\footnotetext[5]{https://github.com/dmis-lab/biobert-pytorch}
\footnotetext[6]{https://github.com/antct/cil}
\footnotetext[7]{https://github.com/MatNLP/HiCLRE}
\footnotetext[8]{https://github.com/dair-iitd/DSRE}

\revised{ Figure \ref{fig:PR_compare} summarizes the experimental
  results.  For convenient observation and fair comparison, we divide
  the above baseline models and the presentation of the results into
  two groups, based on whether BERT is used as the encoder or not,
  i.e. GBRE and non-BERT models in one group GBRE-BERT and BERT-based
  models in the second group.  Tables
  \ref{tab:AUC_compare_biorel}-\ref{tab:AUC_compare_TBGA_bert} report
  various P@N values, Mean P@N values, and AUC values of different
  baseline models and the proposed GBRE based approaches on the BioRel
  and TBGA datasets.  }

\revised{
From Figure \ref{fig:PR_compare}, Table \ref{tab:AUC_compare_biorel} and Table \ref{tab:AUC_compare_TBGA},
our observations about non-BERT models can be summarized as follows:
}
\par

(1) The proposed GBRE method has the best performance for noise reduction.
GBRE significantly improves the performance \revised{over its architectural baseline model} PCNN+ATT on both datasets as shown in Table \ref{tab:AUC_compare_biorel} and Table \ref{tab:AUC_compare_TBGA}. 
\revised{
It outperforms PCNN+ATT by 13.0\% and 14.9\% on BioRel and TBGA, respectively.
Meanwhile, GBRE also outperforms other PCNN+ATT variants, i.e. Intra–Inter Bag and MultiCast,
by at least 6.8\% and 14.7\% on BioRel and TBGA, respectively.
}
Besides, GBRE also achieves the best AUC over both datasets.

(2) GBRE shows high effectiveness in exploiting and using the sentence bag information. 
Compared with the sentence aggregation strategy AVE, which is currently more effective than other strategies over biomedical DSRE datasets, the proposed method adopts ATT as the sentence aggregation strategy but achieves better (higher) AUC over both datasets. 
For instance, our methods outperforms PCNN+AVE by \revised{9.0\%} on BioRel, and BiGRU+AVE by \revised{11.6\%} and TBGA. The proposed GBRE method outperforms BERE+ATT by \revised{10.8\%} on TBGA.
\revised{
Furthermore, compared with MultiCast, which coordinates adversarial training and virtual adversarial training at different levels to boost data utilization, the performance improvement of GBRE on both datasets also demonstrates the significance of better modeling and utilizing the sentence bag information.
}
\par

(3) GBRE always achieves best precision-recall performance compared to all the baselines. 
As observed in Figure \ref{fig:PR_compare}(c), when recall is greater than 0.4,
GBRE is the only method that achieves precision values greater than 0.5,
even BERE+ATT, which fully exploits both semantic and syntactic aspects of sentence information, does not achieve this milestone.
\revised{
From Figure \ref{fig:PR_compare}(a) and (c)
}, it can be seen that the proposed method provides much better precision-recall performance over both datasets than any other baseline models. 

\revised{
From Figure \ref{fig:PR_compare}, Table \ref{tab:AUC_compare_biorel_bert} and Table \ref{tab:AUC_compare_TBGA_bert},
our observations about BERT-based models can be summarized as follows:
}

\revised{
(4) The proposed GBRE method reliably enhances the noise reduction ability of models.
When using BERT as the encoder layer, GBRE improves the performance of vanilla models BERT+ATT and BERT+AVE by a large margin.
As observed in Table \ref{tab:AUC_compare_biorel_bert} and Table \ref{tab:AUC_compare_TBGA_bert},
GBRE-BERT+ATT outperforms BERT+ATT by 10.8\% and 5.0\%, while GBRE-BERT+AVE outperforms BERT+AVE by 2.4\% and 3.8\% on BioRel and TBGA, respectively.
This demonstrates the significance of GBRE as a new approach that consistently provides gains in RE performance for the baseline models.
}

\revised{
(5) The BERT-based GBRE methods achieve significant improvements over other state-of-the-art transformer architecture models.
As observed in Table \ref{tab:AUC_compare_biorel_bert} and Table \ref{tab:AUC_compare_TBGA_bert},
GBRE-BERT+ATT outperforms CIL, PARE and HiCLRE by at least 2.5\% and 8.9\% on BioRel and TBGA, respectively.
Furthermore, compared with PARE and contrastive learning methods, which attempt to utilize the available bag data to the fullest,
GBRE shows greater effectiveness in sentence bag information utilization.
We believe that GBRE is able to provide this performance gain by learning the relevance between sentences and by utilizing inter-sentence level information. 
Through sentence bag self-attention, each sentence gathers and summarizes information from all its immediate neighbor sentences within a bag, according to the degree of relevance between sentences. Thus, GBRE can explore and utilize inter-sentence level information for a sentence bag.
}
 
The comparative results of the above experiments demonstrate the effectiveness and excellent prospects of the proposed method for biomedical relation extraction.

\subsection{\revised{NYT Dataset Evaluation}}
\revised{To further evaluate the effectiveness of the proposed GBRE method, in this section, 
we compare our model with several state-of-the-art methods on NYT-10, the most widely utilized DSRE dataset for general text mining~\cite{2021multicast}.
NYT-10 consists of 570,088 instances, 291,669 entity pairs, and 19,429 relation facts for training and 172,448 instances, 96,678 entity pairs, and 1,950 relation facts for testing.
}

\revised{
In our experiment, we compare GBRE with several competitive baselines, 
including six vanilla models PCNN+ATT~\cite{2016Neural}, Intra–Inter Bag~\cite{2019batt}, MultiCast~\cite{2021multicast},
CIL~\cite{2021cil}, HiCLRE~\cite{2022hiclre}, and PARE~\cite{2022pare},
and two state-of-the-art DSRE models:
\begin{itemize}[leftmargin=*]
\item \textbf{HNRE}~\cite{2018hatt}: a PCNN-based model with hierarchical attention;
\item \textbf{RESIDE}~\cite{2018RESIDE}: a DSRE model integrating the external information including relation alias and entity type.
\end{itemize}
}
\revised{
Note that the results of all baseline models are obtained using the official source codes.
In addition to the already mentioned code repositories,
we used those for HNRE\footnotemark[9] and RESIDE\footnotemark[10].
}
\footnotetext[9]{https://github.com/thunlp/HNRE}
\footnotetext[10]{https://github.com/malllabiisc/RESIDE}

\begin{figure}[ht]
        \centering
        \scriptsize
            \includegraphics[width=0.47\textwidth]{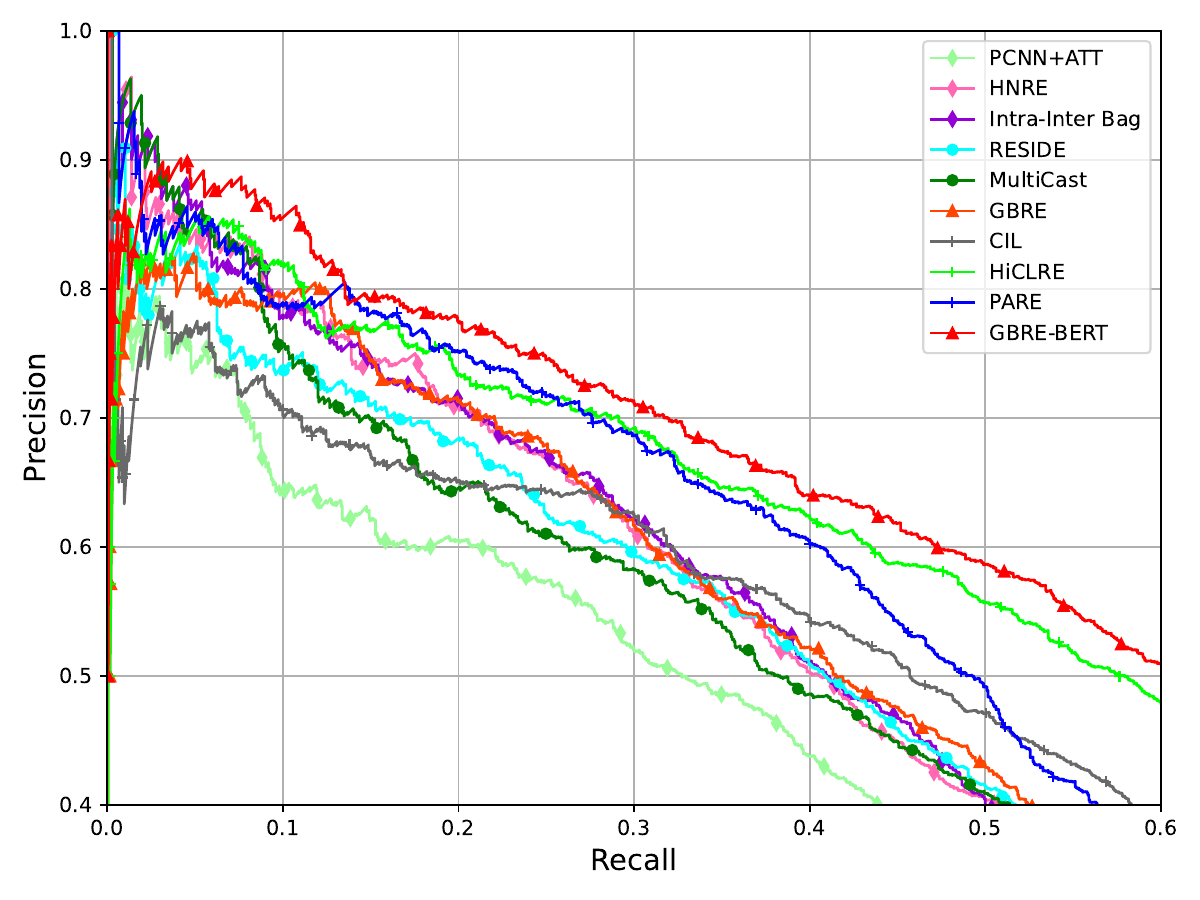}
        \caption{
        \revised{PR curves over NYT-10 dataset for the proposed GBRE model and for several prior methods. 
        Note that GBRE-BERT indicates BERT-based GBRE variant.}
        }
        \label{fig:nyt_pr}
\end{figure}

\revised{
Figure \ref{fig:nyt_pr} shows the PR curves of the proposed model GBRE and the competitors on NYT-10 dataset, 
and Table \ref{tab:AUC_compare_nyt} reports P@N values, Mean P@N values, and AUC values on NYT-10 dataset.
}

\revised{
  From Figure \ref{fig:nyt_pr} and Table \ref{tab:AUC_compare_nyt}, our observations can be summarized as follows.
The proposed GBRE method performs better than the other DSRE methods for noise reduction.
As observed in Table \ref{tab:AUC_compare_nyt}, GBRE outperforms non-BERT models, i.e. HNRE, Intra-Inter Bag, RESIDE and MultiCast, by 0.8\%, 0.7\%, 1.1\% and 1.3\%. 
And BERT-based GBRE method outperforms other Transformer architecture models, i.e. CIL, PARE and HiCLRE, by 9.8\%, 6.2\%, and 2.7\%.
Besides, GBRE-BERT performs best on all of the evaluation metrics. 
Precision-recall curves in Figure \ref{fig:nyt_pr} show that GBRE methods convincingly outperform other models.
}

\revised{
The comparative results of the experiments on NYT-10 dataset demonstrate the effectiveness and universality of the proposed GBRE method in general DSRE domain.
}

\begin{table}[ht]
    \centering
    \caption{
    \revised{
    Performance metrics (\%) for the different RE models on NYT-10 dataset.
    GBRE-BERT indicates BERT-based GBRE variant.
    Models marked with * are BERT-based models, and unmarked models are non-BERT models. 
    The best results among BERT-based models are in bold, and the best results among non-BERT models are \underline{underlined}.
    }
    }
    \begin{tabular}{cccccc}
    \toprule
        {\textbf{Model}} & \textbf{P@100} & \textbf{P@200} & \textbf{P@300} & \textbf{Mean} & \textbf{AUC}\ ($\uparrow$) \\
    \midrule
        PCNN+ATT & 76.0 & 72.5 & 64.0 & 70.8 & 36.3 \\
        HNRE & \underline{85.0} & 81.5 & 77.5 & 81.3 & 41.9 \\
        Intra-Inter Bag & 84.0 & 82.5 & 79.0 & \underline{81.8} & 42.0 \\
        RESIDE & 81.4  & 74.9 & 73.8 & 76.7 & 41.6 \\
        MultiCast & 84.0 & \underline{82.0} & 73.7 & 79.9 & 41.4 \\
        GBRE & 80.0 & 79.0 & \underline{80.0} & 79.7 & \underline{42.7} \\
    \midrule
        CIL* & 76.0 & 72.5 & 70.3 & 72.9 & 44.0 \\
        PARE* & 86.0 & 80.5 & 79.0 & 81.8 & 47.6 \\
        HiCLRE* & 85.0 & 82.5 & 78.0 & 81.8 & 51.1 \\
        GBRE-BERT* & \textbf{89.0} & \textbf{87.0} & \textbf{82.0} & \textbf{86.0} & \textbf{53.8} \\
    \bottomrule
    \end{tabular}
    \label{tab:AUC_compare_nyt}
\end{table}

\subsection{Ablation Experiments}

\begin{figure*}[ht]
        \centering
        \scriptsize
        \begin{tabular}{cc}
             \includegraphics[width=0.47\textwidth]{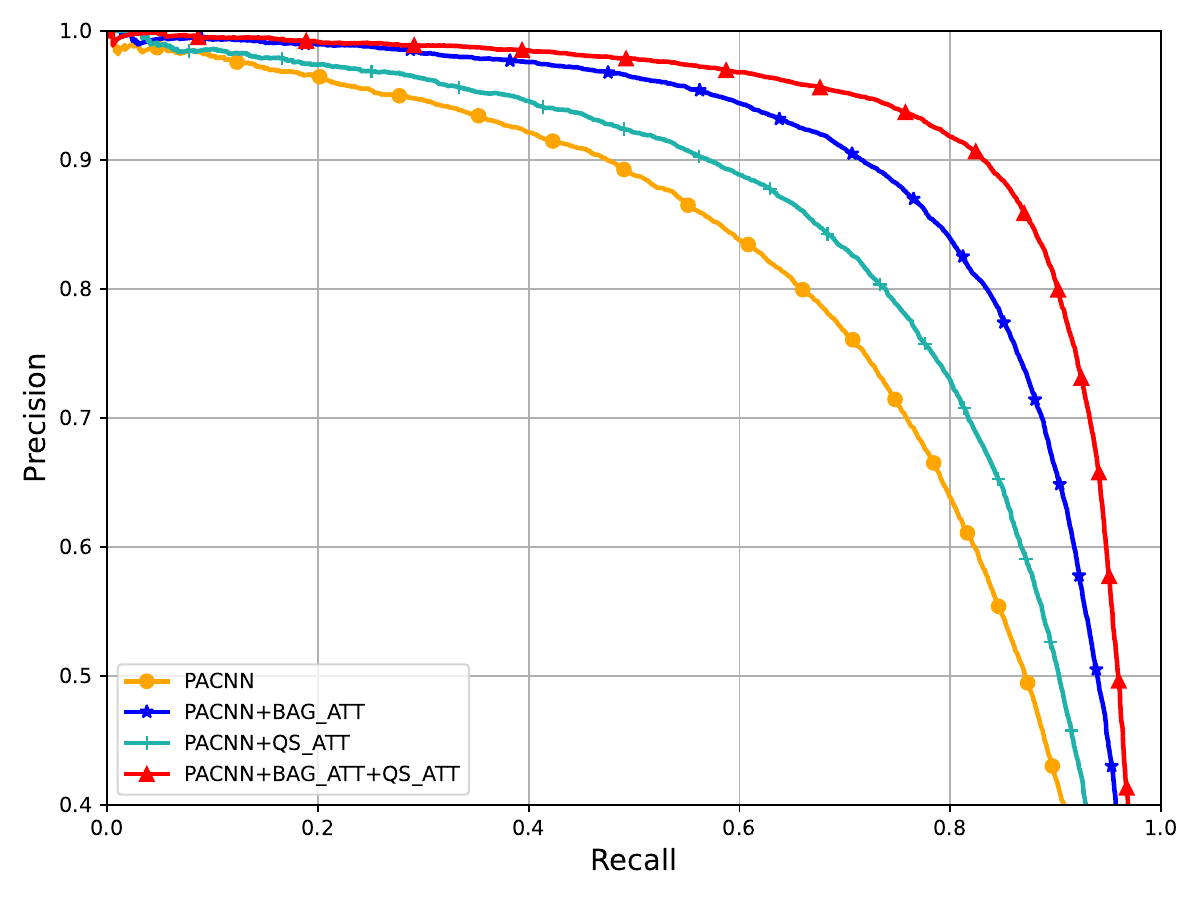} & 
             \includegraphics[width=0.47\textwidth]{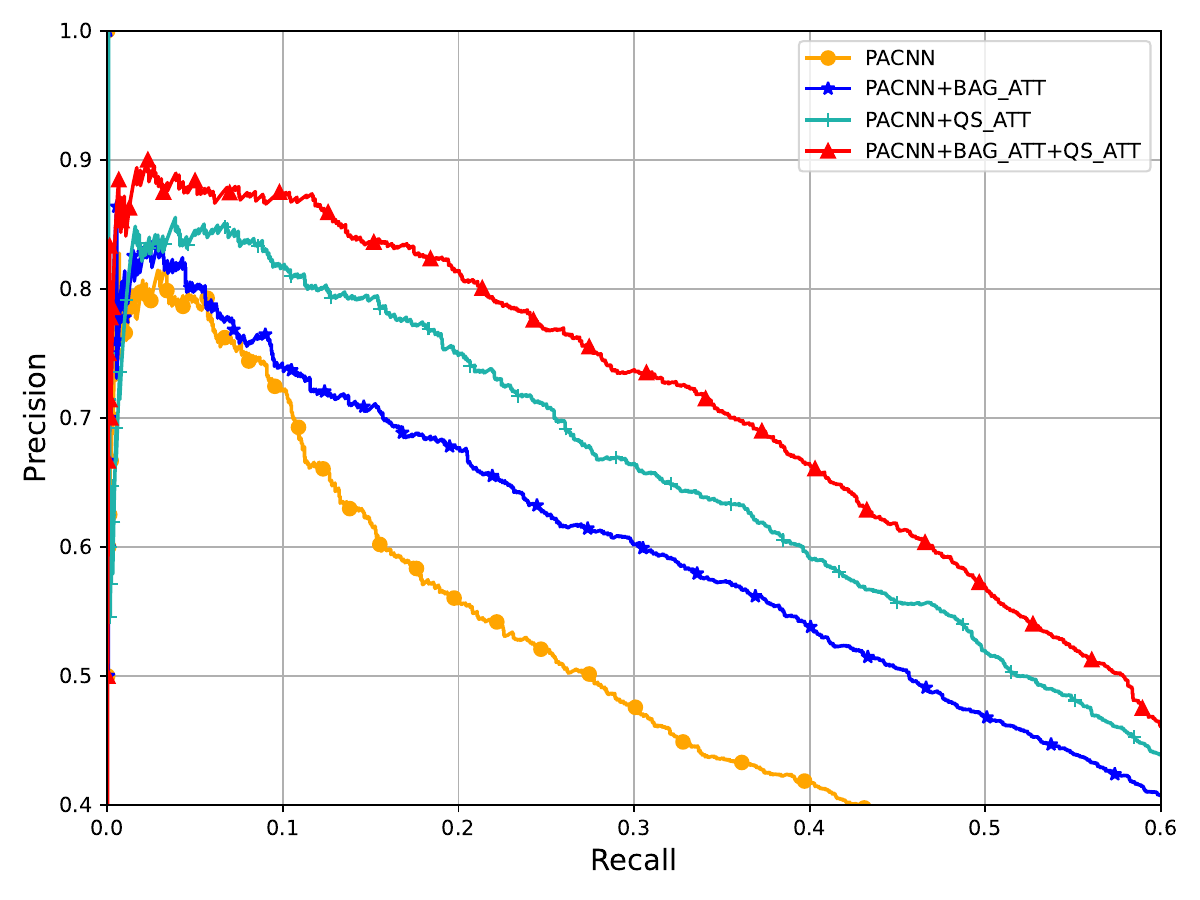} \\
             (a) BioRel & (b) TBGA 
        \end{tabular}
        \caption{PR curves of model with different components on BioRel and TBGA, where PACNN denotes PCNN+ATT, BAG\_ATT denotes sentence bag self-attention, and QS\_ATT denotes query-sentence attention.}
        \label{fig:ablation_pr}
\end{figure*}

\begin{table*}[ht]
        \centering
        \caption{P@N and AUC of model with different components on BioRel and TBGA(\%), where PACNN denotes PCNN+ATT, BAG\_ATT denotes sentence bag self-attention, and QS\_ATT denotes query-sentence attention.}
        \setlength{\tabcolsep}{0.65mm}{
        \begin{tabular}{cccccccccccccc}
            \toprule
            \multirow{2}*{Model} & \multicolumn{6}{c}{BioRel} & \multicolumn{7}{c}{TBGA} \\
            \cmidrule(lr){2-7} \cmidrule(lr){8-14}
            & {P@4000} & {P@8000} & {P@12000} & {P@16000} & {Mean} & {AUC}$( \uparrow )$  & {P@50} & {P@100} & {P@250} & {P@500} & {P@1000} & {Mean} & {AUC}$( \uparrow )$  \\
            \midrule
            PACNN & 96.1 & 91.1 & 83.3 & 73.4 & 86.0 & 79.0 & 76.0 & 75.0 & 74.4 & 70.0 & 62.8 & 71.6 & 40.4 \\
            PACNN+BAG\_ATT & 98.8 & 96.7 & 91.5 & 81.6 & 92.1 & 88.6 & 80.0 & 83.0 & 78.8 & 73.6 & 67.8 & 76.6 & 47.2 \\
            PACNN+QS\_ATT & 96.9 & 93.0 & 86.6 & 76.7 & 88.3 & 83.0 & 80.0 & 84.0 & 84.4 & 80.0 & 73.6 & 80.4 & 51.4 \\
            PACNN+QS\_ATT+BAG\_ATT & \textbf{99.0} & \textbf{97.8} & \textbf{94.9} & \textbf{86.5} & \textbf{94.6} & \textbf{92.0} & \textbf{86.0} & \textbf{89.0} & \textbf{86.8} & \textbf{86.2} & \textbf{78.7} & \textbf{85.3} & \textbf{55.3} \\
            \bottomrule
        \end{tabular}}
        \label{tab:ablation}
    \end{table*}
    
To analyze the contributions and effects of different components of our model, we conducted ablation experiments. We experiment with the base model PACNN and three combinations of our proposed components, PACNN+BAG\_ATT, PACNN+QS\_ATT and PACNN+BAG\_ATT+QS\_ATT. The results from these ablation experiments are shown in Figure \ref{fig:ablation_pr} and  Table \ref{tab:ablation}, based on which, our observations can be summarized as follows:\par

(1) Compared with the base PACNN model, each combination shows significant performance improvements for noise reduction in DSRE. When BAG\_ATT and QS\_ATT are both adopted, we obtain the best results on BioRel and TBGA.\par

(2) PACNN+QS\_ATT always outperforms the base model PACNN regardless of the datasets, and its performance  improvement is relatively stable on BioRel and TBGA. Figure \ref{fig:ablation_pr}(b) shows a large advantage for QS\_ATT over the TBGA on dataset, for which training data contains more noise. 
The results suggest that the QS\_ATT can better capture the critical words in the sentence and effectively reduce sentence noise. \par

(3) PACNN+BAG\_ATT shows large improvements for all evaluation metrics, especially for BioRel which has a large average number of instances per bag. 
\revised{On} TBGA, which has much fewer instances per bag, PACNN+BAG\_ATT still achieves better performance than the model without BAG\_ATT. 
We speculate the reason for the performance gain is that BAG\_ATT can aggregate the information of neighboring sentences and learn the latent relevance between sentences.
When there are more instances in a sentence bag, BAG\_ATT can learn more inter-sentence level information, and contribute a clear improvement.\par

\subsection{Illustrative Example}

\begin{table*}[h]
    \centering
    \caption{Illustrative example from the BioRel test set highlighting how the proposed GBRE framework effectively integrates information over a sentence bag. The three sentences $s_1, s_2, s_3$ form a sentence bag. For each sentence: valid denotes whether the bag label (``has chemical structure'') is correct, for the proposed GBRE method and for PCNN+ATT, the numerical value indicate the selective attention weight, and for PCNN+ONE, 1 or 0 indicates whether the sentence is selected or not.}
    \begin{tabular}{c|c|c|c|c|c}
        \toprule
        {ID} & {Sentence} & {Valid} & {GBRE (proposed)} & {PCNN+ATT} & {PCNN+ONE} \\
        \midrule
        {$ s_1 $} & \makecell{Calcium hopantenate, which is obtained by substituting the \textbf{beta-alanine} \\of pantothenic acid for gamma-\textbf{amino} butyric acid, is a therapeutic drug \\for mental retardation and cerebrovascular dementia.} & {False} & {0.15} & {0.64} & {1} \\
        \midrule
        {$ s_2 $} & \makecell{Uptake of gaba was inhibited by beta-Guanidinopropionic acid, \textbf{beta-alanine},\\ gamma-amino-beta-hydroxybutyric acid,beta-\textbf{amino}-n-butyric acid,\\3-aminopropanesulphonic acid and taurine.} & {False} & {0.10} & {0.07} & {0} \\
        \midrule
        {$ s_3 $} & \makecell{The presence of the characteristic 4'-phosphopantetheine prosthetic group\\ was indicated by the occurrence of equimolar quantities \\of \textbf{beta-alanine} and taurine in \textbf{amino acid} hydrolysates.} & {True} & {0.75} & {0.29} & {0} \\
        \midrule
        \multicolumn{6}{l}{$ Entity \ Pair: beta-alanine, amino \  acid; \ Relation: has\_chemical\_structure $} \\
        \bottomrule
    \end{tabular}
    \label{tab:case_study}
\end{table*}

An instance selection example from the BioRel test set is listed in Table \ref{tab:case_study}. The bag consists of two noisy sentences $ s_1 $ and $ s_2 $, and one valid sentence $ s_3 $ for the relation label “$ has\_chemical\_structure $”.
The baselines PCNN+ATT and PCNN+ONE both predict incorrect relation labels for the entity pair $ (beta-alanine, amino \  acid) $, while the proposed GBRE model correctly predicts the relation label. \par

The values in the table indicate that the noisy sentence $ s_1 $ is assigned the highest attention score by PCNN+ATT and PCNN+ONE, while the valid sentence $ s_3 $ is assigned much lower scores by PCNN+ATT. For the proposed GBRE model, the valid sentence $ s_3 $ is assigned the highest score and the noisy sentences $ s_1 $ and $ s_2 $ are assigned relatively low scores, demonstrating that our model can effectively select the valid instances from noisy data and adequately utilize the sentence bag information. \par

\begin{table}[h]
    \centering
    \caption{Selective attention scores for the sentences for models with different components.}
    \begin{tabular}{cccc}
         \toprule
         {ID} & {$ s_1 $} & {$ s_2 $} & {$ s_3 $} \\
         \midrule
         {PACNN} & 0.64 & 0.07 & 0.29 \\
         {PACNN+QS\_ATT} & 0.028 & 0.001 & 0.971 \\
         {PACNN+BAG\_ATT} & 0.328 & 0.226 & 0.446 \\
         {PACNN+BAG\_ATT+QS\_ATT} & 0.15 & 0.10 & 0.75 \\
         \midrule
         {Valid} & {False} & {False} & {True} \\
         \bottomrule
    \end{tabular}
    \label{tab:case_2}
\end{table}

Table \ref{tab:case_2} compares the scores allocated by the selective attention mechanism to the three sentences  $ s_1 $, $ s_2 $ and $ s_3 $ in the bag when using models with different components: PACNN, PACNN+QS\_ATT, PACNN+BAG\_ATT and PACNN+BAG\_ATT+QS\_ATT. Based on the tabulated values, we can make the following observations:

(1) Compared with base model PACNN, the noisy sentences $ s_1 $ and $ s_2 $ are assigned much lower scores by PACNN+QS\_ATT, while the valid sentence $ s_3 $ is assigned an extremely high score, which indicates that QS\_ATT could effectively reduce sentence noise and help to select valid instances from the sentence bag.\par

(2) Compared with base model PACNN, PACNN+BAG\_ATT assigns the highest score to the valid sentence $ s_3 $, and noisy sentences $ s_1 $ and $ s_2 $ are assigned relatively high scores. Among the two noisy sentences,  $ s_1 $ is mistaken for a valid sentence by PACNN and PCNN+ONE and is assigned a higher score. We believe that the noisy sentence $ s_1 $ contains valuable background information about the target entity pair $ (beta-alanine,\ amino \  acid) $ and there is a higher correlation between $ s_1 $ and $ s_3 $. It demonstrates that BAG\_ATT can effectively explore and learn the latent information between related sentences. \par

(3) When QS\_ATT and BAG\_ATT are both used, there is a clearer distinction between the scores of valid sentences and those of noisy sentences. Furthermore, noisy sentences are also assigned relatively high scores according to how much background information they can provide about the target entity pair.

\begin{table}[h]
    \caption{Factors $ \alpha_{ij} $ ($ i = $ row,  $ j= $ column) reflecting the inter-sentence contributions in the sentence bag self-attention for the PACNN+BAG\_ATT and PACNN+BAG\_ATT+QS\_ATT models.}
    \centering
    \begin{tabular}{ccccccc}
         \toprule
         \multirow{2}*{ID} & \multicolumn{3}{c}{PACNN+BAG\_ATT} & \multicolumn{3}{c}{PACNN+BAG\_ATT+QS\_ATT} \\
         \cmidrule(lr){2-4}\cmidrule(lr){5-7}
         ~ & {$ s_1 $} & {$ s_2 $} & {$ s_3 $} & {$ s_1 $} & {$ s_2 $} & {$ s_3 $} \\
         \midrule 
         {$ s_1 $} & 0.710 & 0.107 & 0.183 & 0.754 & 0.079 & 0.167 \\
         {$ s_2 $} & 0.291 & 0.449 & 0.260 & 0.253 & 0.516 & 0.231 \\
         {$ s_3 $} & 0.175 & 0.091 & 0.734 & 0.118 & 0.078 & 0.754 \\
         \bottomrule
    \end{tabular}
    \label{tab:case_3}
\end{table}

To further examine the ingredients contributing to the attention scores within BAG\_ATT,  in Table \ref{tab:case_3}, we tabulate the factors $ \alpha_{ij} $ in \revised{Equation} (\ref{eq:bag_att_score})  for the PACNN+BAG\_ATT and PACNN+BAG\_ATT+QS\_ATT models in the $ i $-th row and corresponding $ j $-th column of the table. Our observations from this table can be summarized as follows: \par

(1) The valid sentence $ s_3 $ and the noisy sentence $ s_2 $ have higher attention scores to the sentence $ s_1 $. The sentence $ s_3 $ has the higher attention score to the sentence $ s_1 $. These values show that there is a higher degree of relevance between $ s_1 $ and $ s_3 $ and BAG\_ATT can discover and learn the relevance between sentences.
\par

(2) When QS\_ATT is used, all attention scores $ \alpha_{ii} $ increase, which indicates that the relevance of sentence to itself has increased. Although the attention scores $ (\alpha_{ij},\ i \ne j) $ between each sentence and others decrease, the relative order of attention score magnitude does not change between other sentences and the same sentence. We believe that QS\_ATT can reduce the impact of noisy sentences and enable BAG\_ATT to better learn the relevance between sentences.\par

\section{Conclusion and Future Work}

In this paper, we propose a novel graph-based relation extraction (GBRE) framework and demonstrate its effectiveness for distantly-supervised biomedical relation extraction.
The proposed framework is architected to alleviate the problem of noisy labels in DSRE and to exploit the relevance over sentences within a bag under a multi-instance learning formulation. 
Experiments on two popular large-scale biomedical datasets \revised{and the most widely utilized dataset in the general text mining domain} demonstrated that: 
(1) our framework, which views a sentence bag as a graph, can learn the relevance between sentences within a bag and the inter-sentence level information, via message passing in the graph structure.
(2) the query-sentence attention can capture the key words in a sentence that are critical to the relation between an entity pair and effectively reduce sentence noise.
On \revised{all} datasets, the proposed model significantly outperforms the competitive baselines,
\revised{demonstrating universality for both biomedical and general text mining relation extraction.}\par

Beyond the work presented here, effective approaches for integrating in external information (e.g., entity descriptions,  constraint rules and knowledge graphs) are clearly of future research interest in relation extraction. We plan to explore these in future work to further improve relation extraction in the challenging biomedical data setting.

%and significantly improves the performance of biomedical distant supervision relation extraction over prior techniques

% In this paper, we consider alleviating the noisy labeling problem and exploiting the relevance over sentences within a bag to improve the performance of biomedical distant supervision relation extraction model under multi-instance learning.
% For this motivation, we propose a novel graph-based biomedical relation extraction framework, GBRE.
% Experiments on two popular large-scale biomedical datasets have demonstrated that 
% (1) our framework, which views sentence bag as graph, can learn the relevance between sentences within a bag and the he inter-sentence level information, via message passing in graph structure.
% (2) the query-sentence attention can capture the key words in sentence that are more critical to the relation between entity pair and effectively reduce sentence noise.
% The experimental results show that our model significantly outperforms the competitive baselines on both datasets.\par

% In the future, our work will concentrate on exploring external information (e.g., entity descriptions,  constraint rules and knowledge graphs) and effective approaches to integrate these information to further improve model performance on biomedical relation extraction.

\appendix

In this appendix, we highlight how our proposed
  approaches, for treating the sentence bag as a graph and for using a
  synthesized query to exploit query sentence attention mechanisms,
  can also be advantageously incorporated in an alternative DSRE
  pipeline based on the BioBERT~\cite{2020biobert} pretrained language
  model for the biomedical domain. We refer to resulting DSRE
  variants as GBRE-BERT with additional qualifiers specifying further
  design choices.

BioBERT adopts its architecture from the
  BERT~\cite{2019bert} language model and is trained on a combined
  dataset of general and biomedical domain text corpora. BioBERT uses
  WordPiece tokenization~\cite{Wu:GoogleNeuralTranslation:arXiv2016}
  to convert text to a numerical representation for further
  processing. The tokenized representations for an input sentence and
  the auto-generated query (which is the same as in
  Section~\ref{sec:WordEmbeddingLayer}) are concatenated to obtain
  the representation
  $({s}_{1},{s}_{2},...,{s}_{{e}_{1}},...,{s}_{{e}_{2}},...,{s}_{n},{q}_{1},{q}_{2},...,{q}_{m})$,
  which is used for subsequent model operations and serves as the
  input to BioBERT, where ${s}_{{e}_{1}}$ and ${s}_{{e}_{2}}$ are the
  tokens corresponding to the two entities in the sentence. The
  encoder for the BioBERT pretrained language model maps the input
  into a learned embedding space as a sequence of feature vectors:
  $({h}_{1},{h}_{2},...,{h}_{{e}_{1}},...,{h}_{{e}_{2}},...,{h}_{n+m})$,
  where ${h}_{{e}_{1}}$ and ${h}_{{e}_{2}} \in \mathbb{R}^{d}$ are the
  feature vectors corresponding to the entities ${e}_{1}$ and
  ${e}_{2}$, and $ d $ is the dimensionality of the embedding
  space\footnotemark[11].
  We note that by concatenating the auto-generated query with the sentence,
  query-sentence attention is automatically incorporated into the
  features vectors via the multi-headed self-attention mechanism built
  into BioBERT/BERT and a separate query-sentence attention module is
  unnecessary.
  
\footnotetext[11]{The embeddings obtained from BioBERT are context
    aware in contrast with the context-independent neural word
    embedding layer described in Section~\ref{sec:WordEmbeddingLayer}.
    }

    The sentence encoding $s \in \mathbb{R}^{d}$ is
      obtained from the concatenated the feature vectors for entities
      ${e}_{1}$ and ${e}_{2}$ via a learned linear
      transform. Specifically, the sentence encoding is computed as
\begin{equation}
    s = {W}_{s} [{h}_{{e}_{1}}; {h}_{{e}_{2}}] + {b}_{s} 
    \label{eq:bert_non_linear}
\end{equation}
where ${W}_{s}\in \mathbb{R}^{d\times 2d}$ is a trainable weight
matrix, ${b}_{s}\in \mathbb{R}^{d}$ is a trainable bias vector.  The
BioBERT/BERT models also incorporate a sentence classifier, and the
models output a $d$-dimensional feature vector for each sentence
placed at the front of the sentence embedding indicated by a $[CLS]$
tag. This $[CLS]$ feature vector has also been used for relation
extraction in prior work~\cite{2020biobert}. We use the vector $s$
from~\eqref{eq:bert_non_linear} as the sentence encoding for the
subsequent GBRE-BERT pipeline and the $[CLS]$ feature vector is used
as the sentence encoding for the BioBERT baseline.

The sentence encodings are used in our proposed
  sentence bag attention model as already described in
  Section~\ref{sec:SentBagGraphAttention}. Subsequent stages for the
  GBRE-BERT models also use the corresponding processing workflow
  steps already detailed in our mainline description.

{\bf Implementation Details:} We used
  biobert-base-cased-v1.1 checkpoint for BioBERT initialization, and
  bert-base-uncased checkpoint for BERT initialization in the
  BERT-based GBRE methods. The hyper-parameter settings for the
  BERT-based models for the BioRel and TBGA datasets are listed in
  Table~\ref{tab:bert_parameters}. Additional details can be found in
 ~\cite{2019bert} and~\cite{2020biobert}.

\begin{table}[h]
        \centering
        \caption{Hyper-parameter settings for BERT-based models for BioRel and TBGA.}
        \begin{tabular}{cccc}
        \toprule  % 顶部线
             \multirow{2}*{Component} & \multirow{2}*{Parameters} & \multicolumn{2}{c}{Value} \\
             \cmidrule(lr){3-4}
             & { } & {BioRel} & {TBGA}  \\
        \midrule
             \multirow{2}{*}{\makecell{Query-Sentence\\Attention}} & {word size} & {768} & {768}  \\
             & {output size} & {768} & {768}  \\
        \midrule
             \multirow{2}{*}{\makecell{Sentence\\Encoder}} & {hidden size} & {768} & {768}  \\
             & {output size}  & {768} & {768}  \\
        \midrule
            {\makecell{Sentence Bag\\Self-Attention}} & {dropout rate} & {0.3} & {0.25}  \\
        \midrule
            {Classifier} & {input size} & {768} & {768}  \\
        \midrule
             \multirow{4}{*}{Optimization} & {learning rate} & {1e-5} & {5e-5}  \\
             & {dropout rate} & {0.5} & {0.5}  \\
             & {batch size}  & {8} & {16}  \\
             & { optimizer}  & {AdamW} & {AdamW}  \\
        \bottomrule
        \end{tabular}
        \label{tab:bert_parameters}
    \end{table} 
    
% or


\ifCLASSOPTIONcompsoc
  \section*{Acknowledgments}
\else
  \section*{Acknowledgment}
\fi

% This work is supported by the National Natural Science Foundation of China (No. 62071154, 61976071, 61871020) and the National Key Research and Development Program of China (No. 2016YFC0901902).
This work is supported by the National Natural Science Foundation of China (No. 62071154, 61976071,
\revised{62271036}), \revised{the High Level Innovation Team Construction Project of Beijing Municipal Universities (No. IDHT20190506)} and the National Key Research and Development Program of China (No. 2016YFC0901902).

\ifCLASSOPTIONcaptionsoff
  \newpage
\fi



%\bibliographystyle{IEEEtran}
% \end{thebibliography}

\bibliographystyle{IEEEtran}
\bibliography{IEEEabrv,mylib}

\end{document}